\documentclass[pdflatex,sn-mathphys-num]{sn-jnl}

\usepackage{graphicx}%
\usepackage{multirow}%
\usepackage{amsmath,amssymb,amsfonts}%
\usepackage{amsthm}%
\usepackage{mathrsfs}%
\usepackage[title]{appendix}%
\usepackage{xcolor}%
\usepackage{textcomp}%
\usepackage{manyfoot}%
\usepackage{booktabs}%
\usepackage{algorithm}%
\usepackage{algorithmicx}%
\usepackage{algpseudocode}%
\usepackage{listings}%

\usepackage[utf8]{inputenc} 
\usepackage[T1]{fontenc}    
\usepackage{xcolor}         
\definecolor{citeblue}{rgb}{0.41,0.69,0.84}

\usepackage{url}            
\usepackage{booktabs}       
\usepackage{amsfonts}       
\usepackage{nicefrac}       
\usepackage{microtype}      

\usepackage{optidef}
\usepackage{amsmath}
\usepackage{amssymb}
\usepackage{amsthm}
\usepackage{amssymb}
\usepackage{multirow}
\usepackage{multicol}
\usepackage{graphicx}
\usepackage{subcaption}
\usepackage{fancyhdr}

\usepackage{newfloat}
\usepackage{listings}

\theoremstyle{definition}
\newtheorem{definition}{Definition}[section]
\theoremstyle{plain}

\newcommand{\Hc}{\ensuremath{\mathcal{H}}}
\newcommand{\Lc}{\ensuremath{\mathcal{L}}}
\newcommand{\Nc}{\ensuremath{\mathcal{N}}}

\newcommand{\bq}{\mathbf{q}}
\newcommand{\bp}{\mathbf{p}}
\newcommand{\by}{\mathbf{y}}

\newcommand{\Jc}{\ensuremath{\mathcal{J}}}
\newcommand{\RR}{\ensuremath{\mathbb{R}}}
\DeclareCaptionStyle{ruled}
{labelfont=normalfont,labelsep=colon,strut=off} 
\floatstyle{ruled}
\newfloat{listing}{tb}{lst}{}
\floatname{listing}{Listing}
\theoremstyle{thmstyleone}%
\theoremstyle{thmstyletwo}%

\theoremstyle{thmstylethree}%

\renewcommand{\thetable}{\Roman{table}}
\renewcommand{\thefigure}{\Roman{figure}}

\begin{document}

\addtolength{\textheight}{2cm}

\title[Article Title]{Learning Generalized Hamiltonians Using Fully Symplectic Mappings}


\author*[1]{\fnm{Harsh} \sur{Choudhary}}\email{choudhar@fel.cvut.cz}

\author[1]{\fnm{Vyacheslav} \sur{Kungurtsev}}

\author[2]{\fnm{Chandan} \sur{Gupta}}

\author[3]{\fnm{Melvin} \sur{Leok}}

\author[1,4,5]{\fnm{Georgios} \sur{Korpas}}

\affil[1]{\orgdiv{Department of Computer Science}, \orgname{Czech Technical University}, \orgaddress{\street{13, Charles Square}, \city{Prague}, \postcode{12000}, \country{Czechia}}}

\affil[2] {\orgdiv{Department of Computer Science} \orgname{Indraprasth Institute of Information Technology}, \city{Delhi},\country{India}}

 \affil[3]{\orgdiv{Department of Mathematics}, \orgname{UC San Diego}, \postcode{92093}, \state{CA}, \country{USA}}

 \affil[4]{\orgdiv{Quantum Technologies Group}, \orgname{HSBC}, \postcode{117439}, \country{Singapore}}

 \affil[5]{\orgdiv{Archimedes Research Unit on AI}, \orgname{Athena Research Cente}, \postcode{15125}, \city{Marousi}, \country{Greece}}


\abstract{Hamiltonian Neural Networks (HNNs) integrate physical priors into neural models by learning a system's Hamiltonian, improving generalization and sample efficiency.  Identifying the system Hamiltonian from noisy observations of state variables is a challenging task. Moreover, for simulations to faithfully reflect the long-term behavior of Hamiltonian systems, especially energy conservation, it's essential to use symplectic integrators, which are specifically designed to preserve the system’s geometric structure over time. This fidelity comes at a cost: implicit symplectic integrators are computationally intensive and make backpropagation through the ODE solver non-trivial. However, leveraging the self-adjoint property of symplectic integrators, computing gradients w.r.t. Neural Net parameters reduces to solving adjoint equations in the backward pass. In our work, we explore an alternate method of HNN training under noisy observation of trajectories with our HNN model with a symplectic implicit integrator embedded in the loop. Computationally, a predictor-corrector based ODE solver and fixed point iteration help efficient generation of gradient updates. We showcase the numerical advantage, in experiments, in system identification and energy preservation on a range of non-separable, chaotic systems and the efficient computation and memory complexity of our method.}




\maketitle

\section{Introduction}\label{sec:intro}

The Hamiltonian formalism is standard in mathematical physics for expressing dynamics in many physical systems. Beyond leading to reliable schemes to derive ODE dynamics of the system, there are deep structural, geometric, topological, and analytic properties of the Hamiltonian functions that can be used to understand the physics of the system and its dynamics~\cite{MaRa1999, kang1991hamiltonian}. Hamiltonian systems are described by a single scalar function in the phase space $\Hc : \Omega \subseteq \RR^{2n} \rightarrow \RR$ and the corresponding system for $y \in \Omega$ in Hamilton's equations~\cite{hand1998analytical, SanzSerna2018}.

\begin{equation}\label{eq:Hamilt_ODE}
\begin{aligned}
\dot{y} = \mathcal{J}^{-1}\nabla \Hc(y), \quad \mathcal{J} = \begin{pmatrix}
0 & 1\\
-1 & 0
\end{pmatrix},
\end{aligned}
\end{equation}
Where $\mathcal{J}$ is called the canonical symplectic matrix. Solutions of the system are trajectories $y(t)$ which have the property that $\Hc(y(t))$ remains constant at all 
times. If the Hamiltonian is time-invariant, then the Noether quantity associated with time-translational symmetry is conserved along the flow. This implies many important conservation properties across physical systems of interest.
For a set of canonical coordinates \{$p_i$, $q_i$\} $\in \RR^{2d}$,  an integrator, or numerical simulation method, is said to be symplectic if it preserves the canonical symplectic form, $\omega = \sum_{i=1}^{n} dp_i \wedge dq^i$; see subsequent sections and also~\cite{channell1990symplectic}.

Geometric integrators preserve geometric invariants of the flow, and the use of such integrators have been instrumental in understanding the long-term qualitative properties of many important physical systems \cite{Ruth1983,HaLuWa2006,FeQi2010, maslovskaya2024symplectic, valperga2022learning}. 
As such, by learning the Hamiltonian itself rather than to seek to learn trajectories, we are able to learn qualitative properties of the system without the impossible task of numerical handling the sensitivity of such systems. By contrast Non-geometric numerical methods such as the Euler and explicit Runge--Kutta methods do not generally conserve the Hamiltonian over larger time scales \cite{HaLuWa2006}, and this phenomenon inspired the development of energy-preserving and symplectic integration schemes to circumvent this problem. A result in \cite{GeMa1988} shows that it is not generally possible for a fixed-timestep integrator to simultaneously preserve the energy and the symplectic structure.

In recent years, the field of Physics-Informed Machine Learning has brought significant advances in architecture and learning techniques. By enforcing physical inductive biases while learning the dynamics, through appropriate structural modifications to standard NN training, the learning process requires much fewer samples to accurately fit the data and achieves better out-of-distribution accuracy. This includes the recent development of Hamiltonian Neural Networks \cite{bertalan2019learning}. By defining a Hamiltonian as a Neural Network, a more physical representation of the system is available. Incorporating symplectic integrators, and thus accurate reconstruction of the more complex Hamiltonian systems, however, provides technical challenges namely (i) reconstructing an approximate Hamiltonian function from noisy trajectory observations, (ii) solving the implicit equations required by symplectic integration methods, and (iii) ensuring that the learned Hamiltonian model generalizes to out-of-distribution data that we tackle in this work. Motivated by these technical challenges, the remainder of this paper is structured as follows:

In Section \ref{sec:background}, we provide background on two complementary formulations of dynamical systems: the Lagrangian and Hamiltonian perspectives. Section \ref{s:symp_int} then reviews symplectic integrators and their application to solving general Hamiltonian systems where we discuss the shortcoming of using explicit or semi-implicit methods which introduce a structural bias in Hamiltonian leading to the motivation of our work. In Section \ref{sec:method}, we describe the architecture of our Hamiltonian Neural Network and introduce a fully implicit symplectic integrator with a predictor-corrector scheme in the forward pass. Section \ref{ss:adjoint_sens} describes the backward pass and introduces the adjoint sensitivity equations for Hamiltonian Learning. Finally, Section \ref{sec:numerics} presents numerical experiments. 

\subsection{Contributions}
To distill further, the paper makes the following contributions:
\begin{itemize}
    \item Showing that a Neural Framework can be constructed which learns Generalized Hamiltonians upto some error tolerance from noisy observations of trajectories without having any implicit structural bias(no presumptions on the separability of the Hamiltonian).
    \item Showing that the Hamiltonian Neural Net generalizes to out-of-distribution data and is able to learn the governing hamiltonians for systems with non-linear and chaotic dynamics.
    \item An exact, backpropagation-free gradient evaluation method using the self-adjoint structure of the ODE to compute gradients of the loss with respect to neural network parameters. 
\end{itemize}

\section{Hamiltonian Dynamics: A Background}\label{sec:background}
\vspace{0.5em}
Before diving in into the Hamiltonian formulations we provide a generic definition of a manifold in context of mechanics.

\begin{definition}
    In classical mechanics, a \emph{manifold} $Q$ is a mathematical space that locally resembles Euclidean space and serves as the geometric setting for describing the state of a system. The \emph{configuration space} of a mechanical system is modeled as a smooth manifold $Q$, where each point  $q \in Q$  represents a possible configuration of the system.

Associated with $Q$ are two fundamental constructions:
\begin{itemize}
    \item The \emph{tangent bundle} \( TQ \), whose elements are pairs \( (q, \dot{q}) \), representing positions and velocities. This is the natural domain of the Lagrangian formalism.
    \item The \emph{cotangent bundle} \( T^*Q \), whose elements are pairs \( (q, p) \), representing positions and momenta. This serves as the phase space in Hamiltonian mechanics.
\end{itemize}

These bundles are themselves smooth manifolds of dimension \( 2d \), where \( d = \dim Q \). The dynamics of mechanical systems are described by differential equations defined on these manifolds, respecting their geometric and variational structure.
\end{definition}

There are two equivalent and foundational approaches to formulating classical mechanics, both rooted in variational principles: the Lagrangian and Hamiltonian frameworks. These frameworks describe the evolution of physical systems using generalized coordinates and momenta, rather than relying directly on Newton’s second law. The Lagrangian formalism is based on the principle of stationary action and encodes the dynamics in terms of the difference between kinetic and potential energy. The Hamiltonian formalism reformulates the problem as a system of first-order differential equations on phase space, where the state of the system is expressed through canonical coordinates and conjugate momenta. These approaches systematically reveal conserved quantities through symmetries, as formalized by Noether’s theorem, and establish the mathematical structure underlying the dynamics, including the preservation of symplectic geometry in the Hamiltonian setting. Beyond their classical role, these formulations have become central in the analysis of dynamical systems, numerical methods, and emerging applications in optimization and machine learning.

\subsection{The Lagrangian picture}
Lagrangian mechanics describes the motion in a mechanical system by means of the configuration space. The configuration space of a mechanical system has the structure of a differentiable manifold, on which its group of diffeomorphisms acts. The basic ideas and theorems of Lagrangian mechanics are invariant under this group\cite{arnol2013mathematical}, even if formulated in terms of local coordinates. A lagrangian mechanical system is given by a manifold ("configuration space") and a function on its tangent bundle.("the Lagrangian function"). Every one-parameter group of diffeomorphisms of configuration space defines a conservation law (i.e. first integral of equations of motion). A Newtonian potential system is a particular case of a Lagrangian system where the configuration space is Euclidean and the Lagrangian function is the difference between Kinetic and potential energies. The equations of motion are derived via the variational principle by extremizing the action functional \eqref{eq:action} whose domain is infinite space of functions and the dynamics is defined by the function which extremizes this action in that space\footnote{extremum of the action functional refers to a function at which the first variation vanishes; such a point may correspond to a minimum, maximum, or saddle point}.

Let $Q$ be a $d$-dimensional smooth manifold representing the configuration space of a mechanical system. The tangent bundle $TQ$ of $Q$ consists of all pairs $(q, \dot{q})$, where $q \in Q$ denotes a configuration and $\dot{q} \in T_q Q$ is a tangent vector representing the velocity at $q$. In local coordinates $(q_1, \ldots, q_d)$ on $Q$, the tangent bundle is locally coordinated by $(q_i, \dot{q}_i)$ for $i = 1, \ldots, d$. Lagrangian mechanics is formulated on $TQ$, where the dynamics of the system are governed by a Lagrangian function $L: TQ \to \mathbb{R}$ that typically takes the form:
\[
L(q, \dot{q}) = T(q, \dot{q}) - U(q),
\]
with $T$ the kinetic energy and $U$ the potential energy
and the action functional $S: C^2([t_0,t_1],Q) \to \mathbb{R}$ is defined as:
\begin{equation}\label{eq:action}
\begin{aligned}
S[q] = \int_{t_0}^{t_1} L(t, q(t), \dot{q}(t)) dt.
\end{aligned}
\end{equation}
Then, the dynamics of the system is governed by Hamilton's principle:
$$
\delta S[q] = 0,
$$
for all variations $\delta q(t)$ of $q(t)$ that vanish at the endpoints, i.e., $\delta q(t_0) = \delta q(t_1) = 0$. This yields the Euler-Lagrange equations:
\begin{align}\label{eq:lagrange}
\frac{d}{dt} \left( \frac{\partial L}{\partial \dot{q}_i} \right) - \frac{\partial L}{\partial q_i} = 0, \quad \text{for } i = 1, \ldots, d.
\end{align}
\subsection{The Hamiltonian picture}
Another picture is the Hamiltonian formulation of dynamics, which provides an alternative (but equivalent) description of the system on phase space. In this picture, one works on the cotangent bundle $T^*Q$ (the space of pairs $(q,p)$ of generalized coordinates and conjugate momenta) instead of the tangent bundle $TQ$ of $(q,\dot q)$ used in Lagrangian mechanics. The cotangent bundle $T^*Q$ carries a natural symplectic structure, which is central to Hamiltonian mechanics. Specifically, let $\mathcal{M} = T^*Q$ be a $2d$-dimensional smooth manifold. Then the symplectic form $\omega$ is a closed, non-degenerate 2-form on $\mathcal{M}$, and in canonical coordinates $(q_1, \ldots, q_d, p_1, \ldots, p_d)$ on $T^*Q$, it is given by:
\begin{align}\label{eq:symplecticform}
    \omega = \sum_{i=1}^d dq_i \wedge dp_i.
\end{align}

To transition from the Lagrangian $L: TQ \to \mathbb{R}$ to the Hamiltonian formalism, we first define the conjugate momenta by taking partial derivatives of $L$ with respect to the generalized velocities. In local coordinates $q_i$\footnote{Strictly speaking, \( q_i \) should be written as \( q^i \), since it denotes a component of a vector, while \( p_i \) is a component of a covector. However, for notational simplicity—and because we work primarily with individual components when formulating our loss functionals—we use subscripts for both.} on $Q$ , the $i$-th component of momentum is defined as:

\begin{align*}
    p_i = \frac{\partial L}{\partial \dot{q_i}}, \quad \text{for } i=1,\dots,d 
\end{align*}
This definition induces a map known as the Legendre transform, often denoted $\mathbb{F}L: TQ \to T^*Q$. The Legendre transform sends a point $(q,\dot q)$ in the tangent bundle to a corresponding point $(q,p)$ in the cotangent bundle by pairing velocities with momenta. In coordinates, $\mathbb{F}L(q,\dot q) = (,q,p,)$ with $p_i = \partial L/\partial \dot q_i$\footnote{We assume $L$ is regular, meaning the Hessian matrix $\big(\partial^2 L/\partial \dot q_i \partial \dot q_j\big)$ is nonsingular, so that this transformation is locally invertible. (For hyperregular Lagrangians, $\mathbb{F}L$ is in fact a global diffeomorphism $TQ \cong T^*Q$.) Under this assumption, for each $(q,p)$ there is a unique $\dot q$ such that $p_i = \partial L/\partial \dot q_i$.}.

Given this correspondence, we define the Hamiltonian $H: T^*Q \to \mathbb{R}$ as the Legendre transform of the Lagrangian, i.e. the function whose value is the “energy” obtained by trading the velocity dependence of $L$ for momentum dependence. In formula,
\begin{equation}
H(q, p) = \sum_{i=1}^d p_i \dot{q}_i - L(q, \dot{q}),\label{H_in_terms_of_L}
\end{equation}
where $\dot{q}_i$ is expressed as a function of $(q, p)$ by inverting Legendre transform. Equivalently, the dependence on the velocities on the right-hand side can be eliminated by extremizing with respect to the velocities, which is analogous to the approach adopted in Pontryagin's maximum principle. Given $H$, we can define a unique vector field $X_H$ on $T^*Q$, the Hamiltonian vector field, by the condition
\begin{align}\label{Hamilton_eq_abstract}
    dH = \omega(X_H,\cdot).
\end{align}
The Hamiltonian vector field $X_H$ takes the form 
\begin{align}
    X_H = \sum_{i=1}^{d} \frac{\partial H}{\partial p_i}\frac{\partial }{\partial q_i} - \frac{\partial H}{\partial q_i} \frac{\partial}{\partial p_i}.
\end{align}
The integral curves of $X_H$ are the solutions to Hamilton's equations 
$$
\dot{p}_i = -\frac{\partial H}{\partial q_i}, \quad \dot{q}_i = \frac{\partial H}{\partial p_i}, \quad \text{for } i = 1, \ldots, d.
$$
While we have derived Hamilton's equations using local canonical coordinates, \eqref{Hamilton_eq_abstract} defines a global vector field on $\mathcal{M}$.
These equations describe the flow of the system in phase space and are equivalent to the Euler--Lagrange equations \eqref{eq:lagrange} derived from Hamilton's principle if the Legendre transformation is globally invertible and the Lagrangian $L$ is related to the Hamiltonian $H$ by \eqref{H_in_terms_of_L}.

For completeness, the Hamilton--Jacobi partial differential equation,
$$
\partial_t S + H(q, \partial_q S) = 0,
$$
describes the generating function $S$ for the canonical transformation, that maps $(q(t_0),p(t_0))$ to $(q(t_1),p(t_1))$. This provides an alternative method for solving Hamilton's equations \cite{AbMa1978}, and Jacobi's solution to the Hamilton--Jacobi equation is given in terms of the action functional evaluated along the solution of the Euler--Lagrange equations.

 \section{Symplectic Integrators}\label{s:symp_int}
 \vspace{0.5em}

In this section, we discuss the commonly used symplectic schemes in the context of separable and non-separable Hamiltonians and the motivation behind using a fully implicit scheme for integrating the Hamiltonian dynamics. The key idea is that semi-implicit methods, such as symplectic Euler and St\"{o}rmer–Verlet, rely on staggered updates which, when applied to separable Hamiltonians of the form $\mathcal{H}(q, p) = T(p) + V(q)$, admit closed-form integration of subflows and can therefore be implemented explicitly. These schemes, preserve the symplectic structure and can be interpreted as integrating a nearby perturbed Hamiltonian—yielding accurate long-term behavior despite local approximation error. However, this approach introduces an inherent modeling bias, particularly when applied to non-separable systems, as the numerical integrator effectively assumes separability in the underlying dynamics.

\subsubsection{Semi-Implicit Methods}
Consider a separable Hamiltonian $\mathcal{H}(q,p)=T(p)+V(q)$, where $T$ and $V$ are the kinetic and potential energy, respectively. In this case the symplectic Euler update can be obtained by solving two simpler Hamiltonian sub-systems exactly and composing their flows. The flow $\Phi_T^h$ of the kinetic part $T(p)$ over a time-step $h$ is given by solving 
\[
\dot{q}=\nabla_p T(p),\quad \dot{p}=0.
\]
This yields the explicit update 
\[
q(h)=q(0)+h\nabla_pT(p(0)), \quad p(h)=p(0).
\]
Similarly, the flow $\Phi_V^h$ of the potential part $V(q)$ solves 
\[
\dot{q}=0,\quad \dot{p}=-\nabla_qV(q)
\]
giving the update  
\[
q(h)=q(0), \quad p(h)=p(0)-h\nabla_qV(q(0)).
\]
Composing these two exact sub-flows (a first-order Lie–Trotter splitting) yields the symplectic Euler scheme. For example, one convenient ordering is $\Phi_T^h$ followed by $\Phi_V^h$, which gives the update formula (sometimes called the “$TV$” variant of symplectic Euler) \cite{blanes2024splitting}

\begin{equation}\label{eq:exp_euler_int}
\begin{aligned}
q_{n+1} &= q_n + h\nabla_p T(p_n),\\
p_{n+1} &= p_n - h\nabla_q V(q_{n+1})~,
\end{aligned}
\end{equation}

In this separable setting, each substep is integrated in closed form – there are no nonlinear implicit equations to solve. The scheme is therefore explicit. Moreover, $\Phi_T^h$ and $\Phi_V^h$ are the exact solutions of the split Hamiltonians; thus symplectic Euler integrates $\Hc=T+V$ with no local error beyond that incurred by non-commutativity of the subflows.

To make this precise, recall that the Hamiltonian vector field \( X_{\Hc} \) associated with a Hamiltonian \( \Hc \colon T^*Q \rightarrow \mathbb{R} \) is defined via the identity \( \iota_{X_{\Hc}} \omega = \mathrm{d}\Hc \), where \( \omega \) is the canonical symplectic form on \( T^*Q \). In canonical coordinates \( (q, p) \), this yields
\[
X_{\Hc} = \left( \nabla_p \Hc,\, -\nabla_q \Hc \right)^\top.
\]
For a separable Hamiltonian \( \mathcal{\Hc}(q, p) = T(p) + V(q) \), we define the split Hamiltonian vector fields \( X_T = (\nabla_p T(p),\, 0)^\top \) and \( X_V = (0,\,-\nabla_q V(q))^\top \), corresponding respectively to the kinetic and potential parts.

In fact, if $T$ and $V$ were such that $[X_T, X_V]=0$ (equivalently, $\{T,V\}=0$ in Poisson bracket form), then $\Phi_T^h \circ \Phi_V^h = \Phi_{T+V}^h$ exactly – the composition would recover the exact solution in one step. (This situation occurs only in special cases, e.g., $T\equiv T(p)$ and $V\equiv V(q)$) In general, $[X_T,X_V]\neq0$ so the method has a nonzero local truncation error of $O(h^2)$, corresponding to first-order global accuracy. Nevertheless, the key point is that for separable $\mathcal{H}$ the symplectic Euler integrator can be implemented analytically – it uses the closed-form solutions (“flows”) of $T$ and $V$ and thus is explicit and efficient for such problems.

\subsubsection{Lie–Trotter Splitting and Composition Methods}

The procedure above is an instance of a Lie–Trotter splitting. We split the Hamiltonian vector field $X_{\mathcal{H}}$ into two parts $X_T$ and $X_V$ (which generate the flows of $T$ and $V$, respectively), and approximate the full time-$h$ flow $\Phi_{\mathcal{H}}^h=\exp(hX_{\mathcal{H}})$ by the composition $\Phi_{V}^h\circ\Phi_{T}^h=\exp(hX_V)\exp(hX_T)$. This operator splitting yields exactly the symplectic Euler update given above. Since each $\Phi_{T}^h$ and $\Phi_{V}^h$ is a symplectic map (being the exact flow of a Hamiltonian system), their composition is also symplectic.

Thus, symplectic Euler inherits the symplecticity (phase-volume preservation and a variational/Hamiltonian structure) of the true flow. We note that there are two distinct first-order splittings: $TV$ (kinetic then potential, as written above) and $VT$ (potential then kinetic), corresponding to the two standard “symplectic Euler” variants – one explicit in $q$ and the other explicit in $p$. Both are symplectic and of order $1$; their composition in symmetric order yields the familiar St\"{o}rmer-Verlet (leapfrog) integrator of order $2$. In geometric integration terms, symplectic Euler is “Lie splitting” applied to separable $\mathcal{H}$, while St\"{o}rmer–Verlet is “Strang splitting” (second-order, symmetric) \cite{hairer2006geometric}.

Concretely, for a separable Hamiltonian $\mathcal{H}(q, p) = T(p) + V(q)$, the Störmer–Verlet method may be written in Kick–Drift–Kick (KDK) form as:
\[
\begin{aligned}
    p_{n+1/2} &= p_n - \frac{h}{2} \nabla_q V(q_n), \\
    q_{n+1} &= q_n + h \nabla_p T(p_{n+1/2}), \\
    p_{n+1} &= p_{n+1/2} - \frac{h}{2} \nabla_q V(q_{n+1}).
\end{aligned}
\]
This is a symmetric composition of symplectic Euler steps and is therefore second-order accurate. The associated flow map satisfies:
\[
\Phi_{\text{SV}}^h = \exp\left(\frac{h}{2} X_V\right)\exp(h X_T)\exp\left(\frac{h}{2} X_V\right) = \exp\left(h(X_T + X_V) + \mathcal{O}(h^3)\right),
\]
as shown by the Baker-Campbell-Hausdorff expansion \cite[Section IV.5]{hairer2006geometric}. Thus, the local error is $\mathcal{O}(h^3)$, and the global error is $\mathcal{O}(h^2)$, confirming that SV is second-order accurate.

However, this symmetric splitting implicitly assumes that $\mathcal{H}$ is separable. When applied to a general, non-separable Hamiltonian, the splitting introduces a modeling bias: the numerical trajectory exactly conserves a modified Hamiltonian $\tilde{\mathcal{H}}$ that differs from the true $\mathcal{H}$. This can be interpreted as the method learning a nearby perturbed Hamiltonian of the form:
\[
\tilde{\mathcal{H}} = \mathcal{H} + h^2 \delta \mathcal{H} + \mathcal{O}(h^4),
\]
where $\delta \mathcal{H}$ involves commutators like $[X_T, X_V]$ and higher Lie bracket terms \cite[Section V.2.2]{hairer2006geometric}. This backward error formulation shows that SV approximates the flow of a nearby Hamiltonian system, typically one that is effectively separable even when the true dynamics are not. While this may be acceptable for weakly coupled systems or short-time integration, it introduces a splitting error that accumulates over long times or when modeling highly nonlinear interactions.

Hence, while SV is symplectic, time-reversible, and efficient for separable Hamiltonians, it suffers from a fundamental limitation: it cannot accurately resolve the true flow of general, non-separable systems. This motivates the need for fully-implicit symplectic integrators, which do not rely on a separable decomposition and preserve the structure without incurring splitting bias. We turn to these next.

\subsubsection{Fully Implicit Symplectic Methods}
For systems with stiff dynamics or strongly nonlinear Hamiltonians, the semi-implicit methods may suffer from poor accuracy or numerical instability. In such cases, a fully implicit variant is often used:

\begin{equation}\label{eq:imp_euler_int}
\begin{aligned}
    q_{n+1} &= q_n + h \nabla_p \mathcal{H}(q_{n+1}, p_{n+1}), \\
    p_{n+1} &= p_n - h \nabla_q \mathcal{H}(q_{n+1}, p_{n+1}),
\end{aligned}
\end{equation}

(Please note that here we haven't assumed any inherent separability of $\Hc$) This fully implicit symplectic Euler method requires solving a coupled nonlinear system of equations at each time step to compute $(q_{n+1}, p_{n+1})$. Typically, this is done using iterative solvers such as Newton-Raphson. While computationally more expensive, implicit methods provide better stability properties, particularly for stiff or long-time integrations where preserving qualitative behavior is crucial.

In general, fully symplectic integrators are implicit. This is backed by theory: beyond trivial first-order cases, any general symplectic Runge-Kutta scheme must be implicit. 

As we discussed in the previous two subsections, for non-separable Hamiltonian systems,  where $\mathcal{H}$ cannot be cleanly split into independent $T(p)$ and $V(q)$ parts, splitting-based schemes introduce a splitting error. In these cases, a semi-implicit integrator fails to maintain its usual order and symplecticity because the coupled $p$–$q$ updates can no longer be separated. In fact, a method like St\"{o}rmer-Verlet degrades to first-order accuracy when naively applied to a non-separable $\mathcal{H}$. Thus, fully implicit schemes are required to integrate non-separable dynamics while preserving the symplectic structure. 

A prime example is the implicit midpoint method as below(a Gauss–Legendre collocation scheme)
\begin{align*}
    y_{n+1} = y_n + h\Jc^{-1}\nabla\Hc\left(\frac{y_{n} + y_{n+1}}{2}\right)\\ 
\end{align*}
which is second-order accurate and symplectic for general Hamiltonians\cite{channell1990symplectic}. The implicit midpoint rule treats the position and momentum updates in a coupled manner (evaluating forces at the midpoint of the step), and as a result it preserves volume and energy behavior much better than explicit integrators in complex systems.

The implicit midpoint update is a symplectic integrator and it exactly preserves the canonical symplectic 2-form $\omega = \mathrm{d}q\wedge \mathrm{d}p$ (hence volume in phase space) and is a canonical transformation at each step. Equivalently, it conserves all quadratic invariants of the continuous system \cite{hairer2006geometric} and thus is symplectic. Unlike Lie–Trotter or Strang splitting methods, the implicit midpoint integrator does not require splitting the Hamiltonian into subcomponents. It advances the system by treating the full Hamiltonian’s effects simultaneously (in a fully coupled manner) rather than sequentially. As a result, there is no splitting error at all. The only approximation error is the usual truncation error of order $O(h^3)$ per step, with no additional error terms stemming from operator splitting. 

Splitting integrators approximate the true time-$h$ flow $\Phi_\Hc^h = e^{h\Hc}$ (where $\Hc$ is the Hamiltonian vector field) by a composition of sub-flows (e.g. $e^{h\hat A}e^{h\hat B}$ for $\Hc=A+B$ in Lie–Trotter, or $e^{\frac{h}{2}\hat A}e^{h\hat B}e^{\frac{h}{2}\hat A}$ in Strang splitting). This introduces extra error terms due to the noncommutativity of $\hat A$ and $\hat B$ (e.g. leading to an $O(h^2)$ Lie bracket term in the local error). The implicit midpoint rule avoids this entirely by computing a single self-consistent midpoint state $(q_{n+1/2},p_{n+1/2})$ that accounts for the influence of all parts of $\Hc$ over the interval $[t_n,t_{n+1}]$. In practical terms, this means one does not need to interleave sub-steps for kinetic and potential energy or other splits; a single implicit solve captures their combined effect. Thus, implicit midpoint achieves second-order accuracy without the error terms and inconsistencies that splitting can introduce.

\subsection{Constructing Symplectic integrators using Partition Runge-Kutta (PRK) Methods}\label{ss:prk}
\vspace{0.5em}
Partitioned Runge-Kutta (PRK) methods are a class of numerical integrators that are particularly well-suited for systems where the state can be naturally split into multiple components, such as position and momentum in Hamiltonian systems as given by: 
\begin{equation}\label{eq:prk_dyn}
\dfrac{dq}{dt}  = \frac{\partial\Hc(q,p,t)}{\partial p}, \quad \dfrac{dp}{dt}  = -\frac{\partial\Hc(q,p,t)}{\partial q}.
\end{equation}
Now generally, when we are solving for systems, we can combine these equations in a single vector of state $y(t)$ and iteratively solve for $y_n$ using RK methods, however, for Hamiltonian systems these variables play different roles and may evolve at different rates hence these methods are designed to generate separate evolution equations for different variables. Now, equation \eqref{eq:prk_dyn} can be integrated using a partitioned Runge--Kutta scheme:
\begin{align*}
q_{n+1} &= q_n + h_n \sum_{i=1}^{s} b_i k_{n,i},\\ \qquad p_{n+1} &= p_n + h_n \sum_{i=1}^s B_i l_{n,i},
\end{align*}

where
\begin{align*}
k_{n,i} = f(Q_{n,i}, P_{n, i}, t_n + c_ih_n),
\\ \\
\qquad l_{n,i} = g(Q_{n,i}, P_{n,i}, t_n + C_ih_n).
\end{align*}
which are evaluated at the internal stages,
\begin{equation}
\begin{aligned}
Q_{n,i} = q_n + h_n \sum_{j=1}^{s} a_{ij}k_{n,j}, \qquad P_{n,i} = p_n + h_n \sum_{j=1}^s A_{ij} l_{nj}. 
\end{aligned}
\end{equation}
A partitioned Runge--Kutta scheme is symplectic if the following conditions hold:
\begin{equation}
\begin{aligned}
c_i = C_i, 
b_i = B_i, \qquad i &= 1, ..., s; \\ 
b_i A_{ij} + B_j a_{ji} - b_i B_j = 0, \quad i,j&=1,...,s. \\
\end{aligned}
\end{equation}
All the symplectic integrators discussed above, including the one used in our subsequent analysis can be easily constructed using this Symplectic PRK scheme.

\subsection{Symplectic Integration in NNs: Previous Work}\label{ss:prev_work}
\vspace{0.5em}

Since the original proposal by \cite{greydanus2019hamiltonian} and the concurrent work by \cite{bertalan2019learning}, HNNs have generated much scientific interest. This has spawned generative 
\cite{toth2019hamiltonian}, recurrent 
\cite{chen2019symplectic}, and constrained 
\cite{zhong2019symplectic}
versions, as well as Lagrangian Neural Networks 
\cite{cranmer2020lagrangian}
have been proposed.  
\cite{chen2019symplectic} in contrast to standard HNNs, directly optimize the actual states observed at each time step for a given initialization by integrating the partial derivatives using a symplectic integrator (leapfrog algorithm) and backpropagating each squared error through time. The state at the next time step is predicted using the symplectic integrator; in this way the entire time series is predicted, which is then compared with observed states. Note that they make the assumption that the Hamiltonian is separable, which is significant as the leapfrog algorithm is generally implicit, but if the Hamiltonian is separable, then the algorithm becomes explicit.

Most of these works considered either explicit integration schemes, some semi-implicit schemes, or used a separable ansatz for the neural nets. \cite{khoo2024separable, Wu2024} introduced additive separability biases in the HNN architecture/training, allowing the network to learn $H(q,p)=T(p)+V(q)+\text{const}$ forms. This yields better performance by making the Störmer–Verlet (leapfrog) integrators fully explicit and easier to train, as leapfrog-like methods try to approximate a perturbed separable Hamiltonian close to the original one. The error bounds for such symplectic HNNs, for noiseless case have been analyzed in recent works (e.g., showing that energy error grows linearly in time under a symplectic integrator) \cite{canizares2024hamiltonian, david2023symplectic}, reinforcing that the learned $H$ is valid but slightly “off” from the true $H$. These limitations highlight that while semi-implicit methods make training feasible, they come at the cost of a small modeling bias where the integrator inherently treats the Hamiltonian as coming from a separable system to ease computation. More details can be found in \cite{hairer2006geometric, yoshida1990construction}

\cite{xiong2020nonseparable} proposes a generalized HNN framework, which can be used for non-separable Hamiltonians where they approximate the original non-separable Hamiltonian by an augmented one, proposed in \cite{tao2016explicit}, with an extended phase space and a tunable parameter $\omega$ which controls the binding between the two copies of the Hamiltonian and model them using NNs. They propose an in-place symplectic integration scheme for the dynamics. The assumption of augmented Hamiltonian leads to increased complexity with regards to adjusting the binding parameters. As before, the use of the algorithm in \cite{tao2016explicit} is motivated by the fact that it is explicit even for non-separable Hamiltonians. As we mentioned earlier, all such Symplectic HNN architectures learn a valid Hamiltonian which is not necessarily the true Hamiltonian of the system under study. In that case, calculating the error bounds is necessary; there have been a few classic works which derive these error bounds for symplectic integrators, notably \cite{hairer2006geometric, david2023symplectic}.

Most existing works have either relied on explicit methods or employed imperfect implicit schemes, often limited to noiseless settings. This is largely due to the challenge of ensuring convergence of solution trajectories under implicit schemes, thereby motivating the integration of fixed-point iteration techniques within the implicit solver. Another challenge is posed by backpropagation through ODE solvers, which scales in memory with the simulation length and number of parameters. \cite{Rahma2024} proposed a backpropagation-free framework using sampled neural networks for Hamiltonian approximation, effective with rich trajectory data but limited in scope. In contrast, the adjoint method offers greater generality and handles sparse supervision.

\subsubsection{Beyond State of the Art – Full Symplecticity}\label{sss:full_symplecticity}
\vspace{0.5em}

The preceding discussion highlights a central tradeoff in the current landscape of Hamiltonian Neural Networks: while explicit or semi-implicit methods allow for scalable training, they inherently restrict the representational capacity of the learned dynamics by assuming either separability or surrogate approximations of the true Hamiltonian system. As a result, they only \emph{partially} preserve the symplectic structure, which may lead to long-term drift, instability, or degradation of physical consistency in the simulated trajectories, particularly in non-separable, stiff, or noisy regimes.

To address this, we propose a framework that enforces \emph{full symplecticity} during both prediction and training. This approach drops the separability constraint and instead solves the fully implicit update equations dictated by a general, possibly non-separable Hamiltonian using a symplectic integrator. Crucially, we integrate this with adjoint-based training to enable gradient computation through implicit solvers without storing full trajectories, allowing us to scale to longer time horizons and more complex dynamics.

Our method preserves the core geometric invariants of Hamiltonian systems : volume preservation and symplectic two-form conservation across all time steps, and does so without relying on structural assumptions about the form of the Hamiltonian. This facilitates learning accurate, physically faithful models even in settings where traditional HNNs or leapfrog-style methods fail or underperform. In the sections that follow, we detail the construction of our architecture, the fixed-point iteration mechanism that ensures convergence of implicit updates, and our adjoint formulation for scalable optimization.

\section{Fully Symplectic Hamiltonian Neural Network}
\label{sec:method}
\begin{figure*}[!ht]
    \centering
    \includegraphics[width=\textwidth]{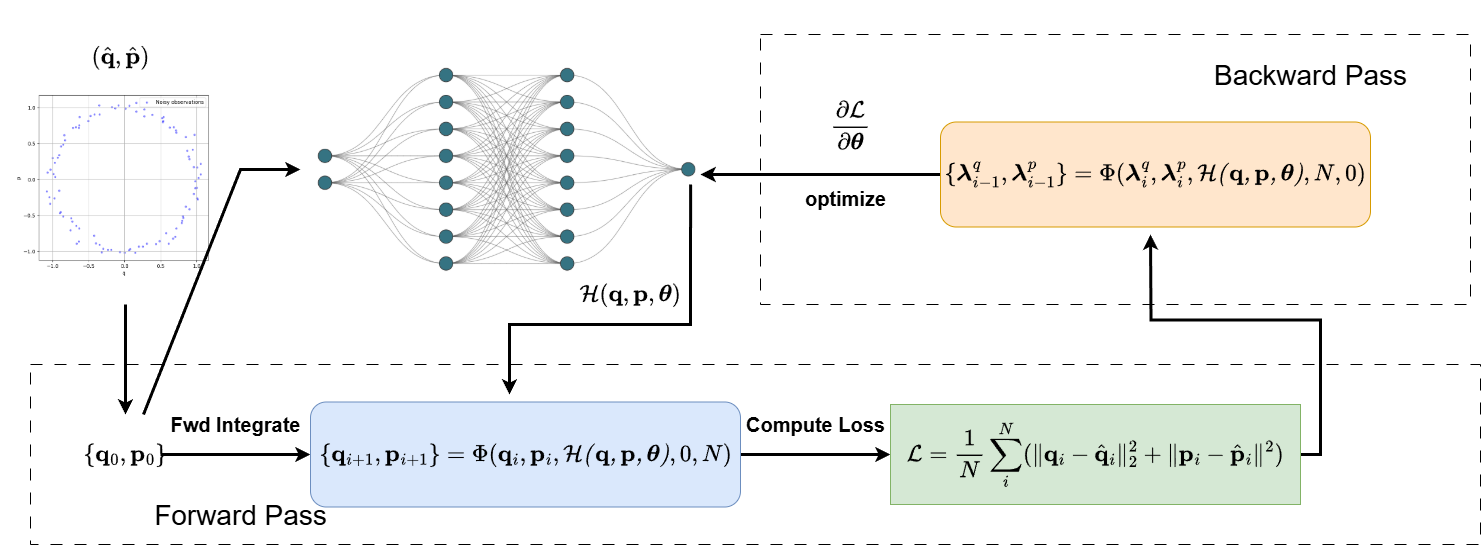}
    \caption{The schematic of the Hamiltonian Identification framework, where the Network represents a parametrized Hamiltonian, the block in the blue below represents the ODE solver in the forward pass, and the block in the orange above represents the same ODE solver but for adjoint dynamics to get the Loss gradients.}
    \label{fig:fig_schematic}
\end{figure*}

\vspace{0.5em}
Now we are at the point where we can apply the insights from symplectic integration theory to the data-driven setting.
Consider the generic problem of learning a Hamiltonian 
$\Hc$ from data, with noisy observations $(\hat{\bq}, \hat{\bp}) \in \RR^{2n}$. We may choose to parametrize it, consider $\Hc(\bq,\bp; \boldsymbol{\theta})$, where $\boldsymbol{\theta} \in \mathbb{R}^{l}$ are the parameters of a neural network and $\bq$, $\bp$ are some coordinates in phase space such that the Hamiltonian is some function of the three arguments. The problem boils down to learning the functional form of Hamiltonian $\Hc(\bq, \bp, \boldsymbol{\theta^*})$ where inference in the network, for given trained $\boldsymbol{\theta}^*$, amounts to computing the value of Hamiltonian function $\Hc(\bq, \bp, \boldsymbol{\theta}^*)$ over phase space provided a value of $(\bq, \bp) \in \RR^{2n}$ with an output depending on $\boldsymbol{\theta}^*$. 
One simulates the Hamiltonian dynamics by computing gradients of $\Hc$ with respect to $\bq$ and $\bp$, then applying a numerical integrator to the first-order dynamics. The application of a symplectic integrator at this stage enforces accuracy and long term stability of the procedure. Algorithm \ref{alg:hamiltonian_identification} describes the complete details of our Hamiltonian learning scheme.
The novelty of this work is that we consider forward and backward passes based on implicit symplectic integrators the schematic of which is shown in the figure \ref{fig:fig_schematic}. This poses a challenge as this typically requires the solution of a nonlinear equation. Fortunately, by the application of a fixed-point iteration, paired with a explicit predictor, this algorithm can converge to the solution quickly.

\begin{algorithm}[H]
\caption{Hamiltonian Identification and Optimization}\label{alg:hamiltonian_identification}
\begin{algorithmic}[1]
\State \textbf{Input:} Initial $\boldsymbol{\theta}^{(0)}$, data $\{(\hat{\bq}_i,\hat{\bp}_i)\}_{i=0}^{N}$, learning rate $\eta$, \texttt{max\_epochs}
\State \textbf{Output:} Optimized parameter $\boldsymbol{\theta}^{*}$

\For{$k = 0$ to \texttt{max\_epochs}}
  \State \textbf{Forward pass}
  \State Define $\Hc(\bq,\bp;\boldsymbol{\theta}^{(k)})$ and access $\nabla_{\bq}\Hc,\nabla_{\bp}\Hc$ as needed
  \State $(\bq_0,\bp_0) \gets (\hat{\bq}_0,\hat{\bp}_0)$
  \For{$i = 0$ to $N-1$}
    \State $(\bq_{i+1},\bp_{i+1}) \gets \Phi\!\big(\bq_i,\bp_i;\Hc(\bq_i,\bp_i;\boldsymbol{\theta}^{(k)})\big)$ \Comment{implicit map \eqref{eq:im_int}, see Alg.~\ref{alg:pc-fpi}}
  \EndFor
  \State $\Lc(\boldsymbol{\theta}^{(k)}) \gets \frac{1}{N}\sum_{i=1}^{N}\!\left(\|\bq_i-\hat{\bq}_i\|_2^2+\|\bp_i-\hat{\bp}_i\|_2^2\right)$ \Comment{Compute loss}

  \State \textbf{Adjoint backward pass}
  \State $\boldsymbol{\lambda}^q_N \gets \partial \Lc / \partial \bq_N$;\quad
         $\boldsymbol{\lambda}^p_N \gets \partial \Lc / \partial \bp_N$ \Comment{Adjoint terminal states}
  \For{$i = N$ to $1$}
    \State $(\boldsymbol{\lambda}^q_{i-1},\boldsymbol{\lambda}^p_{i-1})
      \gets \Phi^{\top}\!\big(\boldsymbol{\lambda}^q_i,\boldsymbol{\lambda}^p_i;\Hc(\bq_i,\bp_i;\boldsymbol{\theta}^{(k)})\big)$ \Comment{implicit map \eqref{eq:im_int}}
  \EndFor
  \State $g \gets \partial \Lc / \partial \boldsymbol{\theta}^{(k)}$ \Comment{Loss gradient \eqref{eq:lossgrad}}

  \State \textbf{Parameter update}
  \State $\boldsymbol{\theta}^{(k+1)} \gets \boldsymbol{\theta}^{(k)} - \eta\, g$
\EndFor
\State \textbf{Return:} $\boldsymbol{\theta}^{*}$
\end{algorithmic}
\end{algorithm}

\footnotetext{Vector-Jacobian products are computed via reverse-mode automatic differentiation, avoiding explicit Hessian construction.}

\subsection{Numerical Forward Pass with Implicit-Midpoint method}\label{ss:int_impl}
\vspace{0.5em}
The Hamiltonian dynamics is a system of ODE given by 

\begin{align*}
\dfrac{dq_i}{dt}  = \frac{\partial \Hc}{\partial p_i}, \quad \dfrac{dp_i}{dt}  = -\frac{\partial \Hc}{\partial q_i}.
\end{align*}

According to the theory of geometric Integration \cite{hairer2006geometric}, symplectic methods should be used to solve these.  The core of the approach lies in applying a predictor-corrector method coupled with fixed-point iterations to handle the non-linear implicit update at each step.
The integrator that we use in our analysis is the Implicit midpoint method which is a fully implicit second-order scheme with updates at each step given by system of equations: 
\begin{equation}\label{eq:im_int}
\begin{aligned}
    q_{i+1} &= q_i + h\nabla_p\Hc\left(\frac{q_{i} + q_{i+1}}{2},\frac{p_{i} + p_{i+1}}{2}, \theta\right)\\
    p_{i+1} &= p_i - h\nabla_q\Hc\left(\frac{q_{i} + q_{i+1}}{2},\frac{p_{i} + p_{i+1}}{2}, \theta\right)
\end{aligned}
\end{equation}
The equation is implicit and thus
the line 8 in Algorithm \ref{alg:hamiltonian_identification} requires fixed point iterations to converge to a good solution. The details of the integrator with fixed-point iteration is discussed in Algorithm \ref{alg:pc-fpi}.

\begin{algorithm}[ht]
\caption{Predictor-Corrector with Fixed-Point Iteration (PC + FPI)}
\label{alg:pc-fpi}
\begin{algorithmic}[1]
\Require Current state $\by_i \equiv (\bq_i, \bp_i)$, step size $h$, dynamics $\nabla_\by \Hc$, number of fixed-point iterations $n$
\State $\by_{{i,0}} \gets \mathrm{RK2}(\by_i)$ \Comment{Predictor}
\For{$j = 1$ to $n$}
    \State $\by_{i,j} \gets \by_{i,0} + h \cdot \nabla_\by \Hc\left(\frac{\by_{i,0} + \by_{i,j-1}}{2}\right)$ \Comment{Implicit Midpoint Update}
\EndFor
\State \Return $\by_{i+1} = \by_{i,n}$
\end{algorithmic}
\end{algorithm}

Starting from an initial point $y_0 \equiv (\bq_0, \bp_0)$, a preliminary estimate for the next step is first obtained using an explicit predictor, followed by refinement through repeated application of the implicit midpoint update via fixed-point iteration. Alternatively, when access to the full trajectory $(\bq_i, \bp_i)$ is available, as is the case in training against noisy trajectories, one may optionally bypass the predictor step and directly use the noisy values as an initial guess. This choice is user-configurable through a command-line argument.

In our implementation, the simulation starts from an initial point 
$\by_0 \equiv \{q_0, p_0\}$. 
Algorithm~\ref{alg:pc-fpi} is first applied with an explicit update to obtain a 
predicted value $y_{\text{pred}}$, which is then used as the initial guess for the fixed-point iterator. 
This iterator runs for $n$ steps and yields 
$\by_1 \equiv (\bq_1, \bp_1)$, which becomes the starting point for computing $y_2$, and so on. 
To evaluate the loss at $N$ points, we perform $N$ forward passes; 
each forward pass involves $n$ fixed-point iterations, 
resulting in a total of $n \cdot N$ function evaluations.

\subsection{Backward pass with gradient computation via Adjoint Sensitivity method}\label{ss:adjoint_sens}
\vspace{0.5em}

Once we forward pass through the ODE solver and formulate the loss function $\mathcal{L}(\bq, \bp)$, we need to optimize with respect to the parameters $\boldsymbol{\theta}$. As discussed in the previous section, pairing the ODE solver with fixed-point iterations incurs multiple function evaluations per time step. To compute gradients for optimization, the automatic differentiation engine must retain the entire sequence of intermediate operations, in order to backpropagate gradients through the solver and evaluate the sensitivity of the loss with respect to $\boldsymbol{\theta}$.
This standard approach, often implemented via reverse-mode automatic differentiation (backpropagation), is memory-intensive, as it requires storing all intermediate values of $(\bq_i, \bp_i)$ and internal solver states—across the entire integration window. This memory overhead grows linearly with the number of time steps and model parameters, and becomes prohibitive for long-horizon simulations or high-dimensional dynamical systems.

To overcome this bottleneck, one can instead formulate the gradient computation as a boundary-value problem via the adjoint sensitivity method. This method derives a backward-in-time differential equation for the adjoint variables (or co-states), which represent the sensitivity of the loss functional with respect to the trajectory. Crucially, the backward pass is decoupled from the original forward pass: the adjoint equations can be solved without retaining the entire forward trajectory. This reformulation allows memory-efficient gradient computation, particularly when the number of scalar outputs (loss terms) is much smaller than the number of parameters.

The sensitivity analysis by the adjoint method has its roots in several fields, such as control theory, geophysics,
seismic imaging, and photonics \cite{jorgensen2007adjoint, xiao2021deep}. The adjoint method gives the gradients of a cost that is in Lagrange form, and this has gained some traction recently in the deep learning community, after it was shown in the Neural ODEs paper 
\cite{chen2018neural}
that it is possible to parameterize the vector field defining an ODE by a neural network and differentiate along the flow to learn the vector field. Since then, the adjoint method has been mostly used in the Neural ODE context as a constant memory gradient computation technique. In the subsequent subsection, we show how the adjoint equations derived in NeuralODE case cannot be used exactly in this scenario. Here in this work we follow the classic ODE constraint optimization-based approach where we assume that our dynamics comes from  a parametrized Hamiltonian $\Hc(\theta, {q}, {p})$ where q and p here are functions of time $t$. The optimization is then w.r.t. the Neural Network parameters which minimize the squared loss. 
\begin{mini}|l|
 {\theta}{\Lc(\boldsymbol{\theta}, q, p) = \frac{1}{T}\sum \limits_{i=0}^T \|q_i - \hat{q}_i\|^2 + \|p_i - \hat{p}_i \|^2}{}{}
\addConstraint{\dot{q}(t, \boldsymbol{\theta}) - \frac{\partial\mathcal{H}(\theta, q, p)}{\partial p}}{=0}
\addConstraint{\dot{p}(t, \boldsymbol{\theta}) + \frac{\partial\mathcal{H}(\boldsymbol{\theta}, q, p)}{\partial q}}{=0}.
\label{eq:opt_prob_1}
\end{mini}

Where $(q_i, p_i)$ is the predicted value of canonical coordinates through the symplectic map $\Phi_h$ at iteration $i$ where $h$ is the stepsize and $(\hat{q}_i, \hat{p}_i)$ is the true value of those variables. For simplicity, here we consider q and p as scalar functions. In a classic case, the objective is a continuous limit of this discreet loss which would read
    \begin{align*}
      \frac{1}{T}\int\limits_0^T \|q(\boldsymbol{\theta}, t) - \hat{q}(t)\|^2 + \|p(\boldsymbol{\theta}, t) - \hat{p}(t)\|^2 dt 
\end{align*}
Now, considering $\Phi_h$ as a general symplectic map and writing a shorthand notation $y_i$ for $(q_i, p_i)$, we can write
\begin{align*}
       \by_i = \Phi_h(\by_i, \by_{i-1})
\end{align*}
In usual cases where a Neural Network is used as a function approximator, the gradients are computed using backpropagation which is a standard way to optimize the cost function adjusting the network weights. In our case, the forward pass includes a Symplectic ODE solver which means we have to backpropagate through this solver which will drastically increase the memory requirements for computational graph which can be understood if we
consider our cost function
\begin{align*}
    \mathcal{L} = \frac{1}{N}\sum\limits_{i=1}^N (y_i - \hat{y}_i)^2,
\end{align*}
where each $y_i$ is obtained via forward propagation through a symplectic ODE solver. It is clear that if we are solving for larger simulation lengths, the number of evaluations in the backward pass will scale with the number of timesteps as shown below:
\begin{align*}
    \frac{d\mathcal{L}}{d\boldsymbol{\theta}} = \sum\limits_{i} \frac{\partial \mathcal{L}}{\partial y_i} \sum\limits_{j=1}^{i-1} \frac{\partial y_i}{\partial y_j} \frac{\partial y_j}{\partial \boldsymbol{\theta}},
\end{align*}
where each $y_i$ is a function of $y_{i}, y_{i-1}, y_{i-2},...y_{1}$.  In contrast to this, if we use adjoint sensitivity analysis, we first have to derive the adjoint equations which is a system of ODEs in the adjoint variable $\lambda(t)$ which is a time dependent version of Lagrange multiplier which get introduced while solving equation \eqref{eq:opt_prob_1} using variational calculus. The system of adjoint ODEs is then given as
\begin{equation}\label{eq:Hamiltonian_system}
    \begin{aligned}
        \frac{d}{dt}
        \begin{pmatrix}
            q(t) \\
            p(t) 
        \end{pmatrix} =
        \begin{pmatrix}
            \nabla_p \Hc(q, p, \boldsymbol{\theta}) \\
            -\nabla_q \Hc(q, p, \boldsymbol{\theta})
        \end{pmatrix},
    \end{aligned}
\end{equation}
where $ q(0) \coloneqq q_0$ and $p(0) \coloneqq p_0$. We further define  
\begin{align*}
    a_{\rm aug} = 
    \begin{pmatrix}
        \nabla_q \Lc \\
        \nabla_p \Lc \\
    \end{pmatrix}, \quad
    f_{\rm aug} =
    \begin{pmatrix}
        \nabla_p \Hc \\
        -\nabla_q \Hc \\
    \end{pmatrix}, \quad
    z_{\rm aug} =
    \begin{pmatrix}
        \nabla_q \Lc \\
        \nabla_p \Lc
    \end{pmatrix}.
\end{align*}
Based on these definitions, the corresponding adjoint system for the Hamiltonian dynamics is given by:
\begin{align*}
\frac{d}{dt} 
a_{\rm aug} 
&= - \frac{\partial f_{\rm aug}}{\partial z_{\rm aug}}a_{\rm aug},
\label{eq:adjoint_hamiltonian_system}
\end{align*}
which when applied to the Hamiltonian system, results in: 
\begin{equation}\label{eq:adj_eqns}
\begin{aligned}
\frac{d}{dt} \begin{pmatrix}
\lambda^q \\
\lambda^p \\
\end{pmatrix}
&= -\begin{pmatrix}
\nabla_{qp} \Hc &  -\nabla_{qq} \Hc \\
\nabla_{pp} \Hc & -\nabla_{pq} \Hc
\end{pmatrix}
\begin{pmatrix}
\lambda^q \\
\lambda^p \\
\end{pmatrix},
\end{aligned}
\end{equation}
where  $\lambda_q = \nabla \Lc_q$, $\lambda_p = \nabla \Lc_p$. This gives:
\begin{equation}\label{eq:adj_bckwd_eqns}
\begin{aligned}
\frac{d}{dt} \begin{pmatrix}
\lambda^q \\
\lambda^p 
\end{pmatrix}
&= \begin{pmatrix}
-\lambda^q\nabla_{pq} \Hc + \lambda^p\nabla_{qq} \Hc \\
-\lambda^q\nabla_{pp} \Hc + \lambda^p\nabla_{pq} \Hc\\
\end{pmatrix}.
\end{aligned}
\end{equation}
Solving these equations backwards in time subject to :
\begin{equation}\label{eq:adj_terminal_condns}
\begin{aligned}
    \lambda^q(T) = \frac{d\Lc}{dq}_{(t=T)}\text{ and } \quad \lambda^p(T) = \frac{d\Lc}{dp}_{(t=T)},
\end{aligned}
\end{equation}
gives the adjoint state which is in-turn used to calculate the loss gradients solving the following integral:
\begin{equation}\label{eq:lossgrad}
     \frac{d\Lc}{d\theta} = \int_0^T \lambda^T \frac{\partial f_{\rm aug}}{\partial \boldsymbol{\theta}} dt.
\end{equation}
Introducing the adjoint sensitivity method for gradient calculation will amount to solving the adjoint equations backward in time which is a constant memory task where we only need to store the current variable and its partial derivatives in the memory at any particular instant. The complete derivation of HNN adjoint state and gradients is provided in subsequent section.

If we notice, to get the adjoint state, we have to solve equation \eqref{eq:adj_bckwd_eqns} subject to the terminal conditions \eqref{eq:adj_terminal_condns} $t=T$ to $t=0$ and store the results $\lambda(t)$ in memory to later solve the integral, however, to truly leverage the adjoint state method's constant memory advantage, we can perform these two steps concurrently where we evaluate the adjoint variable $\lambda(t)$, then instead of storing it in the memory, accumulate it to a variable which calculates the integral using trapezoidal method and accumulates the result in a variable $grad$, we only have to store the current variable in the memory hence, significant reduction in peak memory complexity. 
\subsection{An ODE constraint optimization-based proof of Adjoint Sensitivity}\label{ss:adjoint_proof}
Here we will present a classic proof of adjoint sensitivity analysis rooted in ODE constraint optimization.
Consider the problem:
\begin{mini}|l|
 {y}{F(\theta, y) = \int \limits_0^T f(\theta, y, t)dt}{}{}
\addConstraint{\dot{y} - h(y, \theta)}{=0}
\addConstraint{y(0)}{=y_0}.
\label{eq:opt_prob_adj}
\end{mini}
where $\theta$ is a vector of unknown parameters, $y$ is a (possibly vector-valued) function of time, $h(\dot{y}, y, \theta) = 0$ is an ODE in implicit form, and $g(y(0), \theta) = 0$ is the initial condition, which is a function of some of the unknown parameters. The ODE $h$ may be the result of semi-discretizing a PDE, meaning that the PDE has been discretized in space but not time. An ODE in explicit form appears as $\dot{y} = \bar{h}(y, \theta, t)$, and so the implicit form is $h(\dot{y}, y, \theta) = \dot{x} - \bar{h}(x, p, t)$.

A gradient-based optimization algorithm requires the user to calculate the total derivative (gradient)
\begin{align*}
d_{\theta}F(y, \theta) &= \int_0^T \left[ \frac{\partial f}{\partial y} \frac{dy}{d\theta} + \frac{\partial f}{\partial \theta} \right] dt.
\end{align*}
Calculating $\frac{dy}{d{\theta}}$ is difficult in most cases. There are two common approaches to simplify this process. One approach is to approximate the gradient $d_{\theta}F(x, \theta)$ by finite differences over $\theta$ (Discretize-optimize). Generally, this requires integrating $n_\theta$ additional ODEs. The second method is to develop a second ODE, this one in the adjoint vector $\lambda$, which is instrumental in calculating the gradient(Optimize-Discretize). The benefit of the second approach is that the total work of computing $F$ and its gradient is approximately equivalent to integrating only two ODEs.

The first step is to introduce the Lagrangian corresponding to the optimization problem:
\begin{equation}
\begin{aligned}
L &\equiv \int_0^T \left[ f(y, \theta, t) + \lambda \left(\dot{y} -  h(y, \theta)\right) \right] dt + \mu (y(0) -y_0).
\end{aligned}
\end{equation}
The vector of Lagrangian multipliers $\lambda$ is a function of time, and $\mu$ is another vector of multipliers associated with the initial conditions. 
Let us write this as
\begin{equation}\label{eq:adj_lagr}
\begin{aligned}
  L \equiv F(y(T)) + &\int_0^T \lambda(t) (\dot{y} - h(y, \theta))  dt + \int_0^T\mu(t) g(y(0), \theta).  
\end{aligned}
\end{equation}

Because the two constraints $h = 0$ and $g = 0$ are always satisfied by construction, we are free to set the values of $\lambda$ and $\mu$, and we can say that
\begin{align*}
    \frac{dL}{d\theta} = \frac{dF(y(T))}{d\theta}
\end{align*}
Now, look at the first integral in equation \eqref{eq:adj_lagr}. Using integration by parts, we can write
\begin{align*}
    \int_0^T \lambda(t)(\dot{y}(t) - h) &= \int_0^T \lambda(t)\dot{y}(t) - \int_0^T \lambda(t) hdt\\
    &= \lambda(t) y(t)\biggr\rvert_{{t=0}}^{t=T} - \int_0^T\dot{\lambda}(t)y(t)dt- \int_0^T \lambda(t) hdt\\
    \end{align*}
Hence the above expression can be written as:
\begin{equation}\label{eq:lambda_util}
    \begin{aligned}
      \int_0^T \lambda(t)(\dot{y}(t) - h)&=\lambda(T)y(T) - \lambda(0)y(0)- \int^T_0 (\dot{\lambda}y + \lambda h )  
    \end{aligned}
\end{equation}
    
Taking derivative w.r.t. $\theta$ gives:

    \begin{align*}
        \frac{dL}{d\theta} = \frac{\partial L}{\partial y(T)}\frac{dL}{d\theta} - \frac{d}{d\theta}\left[\int_0^T \lambda(t)(\dot{y}(t) - h)\right]
    \end{align*}

Now using the relation from equation \eqref{eq:lambda_util}, and simplifying, we get
    \begin{align*}
        \frac{dL}{d\theta} &= \left[\frac{\partial L}{\partial y(T)} - \lambda(T)\right]\frac{dy(T)}{d\theta}  + \int_0^T\left(\dot{\lambda}(t) + \lambda(t)\frac{\partial h}{\partial y}\right)\frac{\partial y(t)}{\partial{\theta}}dt + \int_0^T\lambda(t)\frac{\partial h}{\partial \theta}dt 
    \end{align*}

As we claimed above as well, we are free to choose $\lambda(t)$ such that we can evade difficult to calculate derivatives, which are:
\begin{align*}
    \frac{dy}{d\theta}, \quad \frac{dy(T)}{d\theta}, 
\end{align*}
Making the coefficients zero identically, we are left with the following equations
    \begin{align*}
        \dot{\lambda}& = -\lambda(t)\frac{\partial h}{\partial \theta} \quad
        \text{s.t.} \quad \lambda(T) = \frac{\partial L}{\partial y(T)}
    \end{align*}
Once we make the coefficients zero, the derivative of our objective with respect to the parameters become
\begin{equation}\label{eq:loss_grad}
    \begin{aligned}
        \frac{dL}{d\theta} = \int_0^T \lambda(t)\frac{\partial h}{\partial \theta}dt
    \end{aligned}
\end{equation}

Now, we can safely replace 
\[
y = \begin{pmatrix}
            q(t) \\
            p(t) 
        \end{pmatrix}, \qquad
 h = \begin{pmatrix}
     \nabla_p \Hc\\
     -\nabla_q \Hc
 \end{pmatrix} \quad and  \quad \lambda = \begin{pmatrix}
     \lambda^q(t)\\
     \lambda^p(t)
 \end{pmatrix}
\]
and we will end up with \eqref{eq:lossgrad}

\color{black}{}

\subsection{Comparison with Neural ODE adjoint state method}\label{ss:comp_neural_ODE}
\vspace{0.5em}
The Neural ODE work \cite{chen2018neural} derived the adjoint equations as an efficient framework for backpropagation through an ODE solver where they define an adjoint state \textbf{a(t)} which evolves as
\begin{equation}\label{eq:neuroODE_adj_state}
\begin{aligned}
     \frac{da}{dt} = -a(t)\frac{\partial h}{\partial z} 
\end{aligned}  
\end{equation}
and the expression for loss gradient becomes
    \begin{align*}
        \frac{dL}{d\theta} &= \int_{t_0}^{t_f}a(t)\frac{\partial h}{\partial \theta}\\
          \end{align*}
If we write it as:
\begin{align*}
\frac{dL(t)}{d\theta} &= -\int_{t_f}^{t} a(t) \frac{\partial h}{\partial \theta} dt
\end{align*}

Then if we take the derivative w.r.t. both side, it will give us the loss gradients at any arbitrary time 
\begin{align*}
\frac{d}{dt}\left(\frac{dL(t)}{d\theta}\right) &= - a(t) \frac{\partial h}{\partial \theta} dt
\end{align*}
Now the above equation can be combined with equation \eqref{eq:neuroODE_adj_state} to write

    \begin{align*}
        \frac{d}{dt}\begin{pmatrix}
            a(t)\\
            \frac{dL}{d\theta}
        \end{pmatrix}
        = \begin{pmatrix}
            -a(t)\frac{\partial h}{\partial z}\\
            -a(t)\frac{\partial h}{\partial \theta}
        \end{pmatrix}
    \end{align*}

The benefit of doing this is that now we can use batch parallelism and solve these ODEs simultaneously using SIMD vectorization on CPUs or GPU acceleration (CUDA, TensorFlow, JAX, PyTorch), note that these are independent ODEs. However, great care should be taken when applying Neural ODE frameworks to systems requiring symplectic structure preservation as converting an integral formulation into an ODE system introduces numerical inconsistencies, as the resulting discretization may not respect the underlying geometric properties of the original problem. Moreover, this coupling restricts flexibility, particularly in choosing optimal quadrature points or specialized integrators for evaluating the integral. However, when we integrate this separately from the Adjoint ODE, we are free to choose the discretization/quadrature points for our numerical integrator. Symplectic solvers are designed to maintain energy and phase-space structure and this kind of framework might not preserve symplecticity.

\section{Numerical Results}

\label{sec:numerics}
\vspace{0.5em}
In this section, we present the performance of our method on three representative Hamiltonian systems: the Double Well potential, the Coupled Harmonic Oscillator, and the Hénon-Heiles system. These systems are chosen to cover a range of problems, including separable and generalized non-separable Hamiltonians. For training, the data is generated by sampling initial conditions from a uniform random distribution within a specified bounded domain. In contrast, for evaluation, the model is tested on test data sampled from 3 separate distributions different from training data. 


\subsection{Implementation details }
\vspace{0.5em}
For each experiment, we generated 16,384 initial conditions for the training set and 8,192 for the validation set. Using a high-order symplectic solver, trajectories were simulated with a time discretization of $\Delta t = 0.001$. Additive Gaussian noise $\mathcal{N}(0, 1)$ was introduced at each timestep, scaled by a noise level coefficient of $0.01$.
\subsubsection{Hyperparameter details}
A minibatch size of 512 was used for both the training and validation datasets. The Ansatz Neural Network architecture consisted of three hidden layers with widths 16, 32, and 16, respectively, and a single scalar output. The input layer had a dimensionality of $2^d$
 , corresponding to the phase space variables of the system under study. Model parameters were optimized using the Adam optimizer with an initial learning rate of $10^{-3}$
 . A learning rate scheduler based on the ReduceLROnPlateau strategy was employed to adapt the learning rate during training in response to stagnation in validation loss. The HNN was trained for 10 epochs. For each batch, the starting point was sampled at a random timestep along the trajectory, allowing the model to observe a wide variety of initial conditions and improve generalization across the phase space. In the forward pass, we simulated trajectories for i=6 timesteps and evaluated the loss according to equation~\eqref{eq:final_objective}. Gradients were computed using the adjoint state method, employing the same integrator as used in the forward pass to leverage the self-adjoint property of the integrator.
\begin{equation}\label{eq:final_objective}
    \Lc = \frac{1}{N}\sum\limits_{i=1}^{N} \|\bp_i^{pred} - \bp_i^{noisy}\|_2^2 + \|\bq_i^{pred} - \bq_i^{noisy}\|_2^2
\end{equation}
where state variables $(\bq^{pred}_i, \bp^{pred}_i)$ evolve under a symplectic integrator.

\subsubsection{Evaluation Criteria} 
We aim to learn the functional form of the Hamiltonian \( \mathcal{H}(\bq, \bp) \) beyond simple trajectory matching. Unlike prior works such as Hamiltonian Neural Networks (HNN) and Neural ODEs, which primarily assess learned models by comparing predicted and ground-truth trajectories, we explicitly evaluate the learned Hamiltonian function \( \mathcal{H} \) across the broader phase space. To do so, we sample test points \( (\tilde{\bq}_i, \tilde{\bp}_i) \) from three distinct distributions over the phase space:

\begin{itemize}
    \item \textbf{Random Uniform:} Points are sampled independently from a uniform distribution over a fixed bounding box that encompasses the training trajectories. This probes generalization to randomly scattered unseen states.

    \item \textbf{Uniform Square Grid:} A structured grid of evenly spaced points is generated within the same bounding box. This enables a systematic and resolution-controlled evaluation of \( \mathcal{H} \) over the phase space.

    \item \textbf{Gaussian:} Points are sampled from a multivariate Gaussian distribution centered around typical states observed during training.
\end{itemize}

\begin{figure}[!ht]
    \centering
    \begin{subfigure}{0.31\textwidth}
        \includegraphics[width=\textwidth]{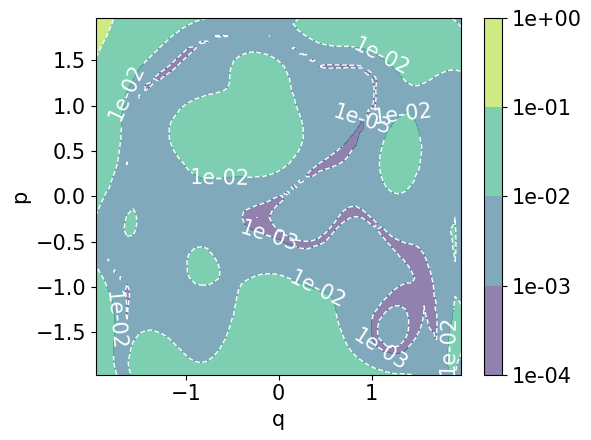}
        \caption{}
        \label{fig:dw_uniform}
    \end{subfigure}
    \hfill
    \begin{subfigure}{0.31\textwidth}
        \includegraphics[width=\textwidth]{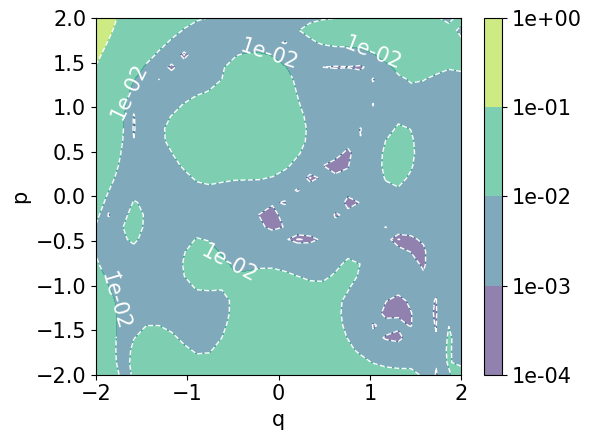}
        \caption{}
        \label{fig:dw_grid}
    \end{subfigure}
    \hfill
    \begin{subfigure}{0.31\textwidth}
        \includegraphics[width=\textwidth]{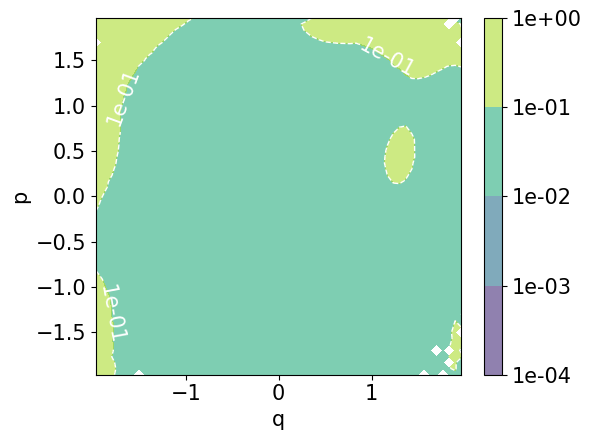}
        \caption{}
        \label{fig:dw_gaussian}
    \end{subfigure}
    \caption{The Hamiltonian prediction error $\|\Hc_{pred} - \Hc_{true}\|_1$ in double well system on test data drawn from 3 different distributions (a)random uniform (b)uniform square grid (c)multivariate gaussian \Nc(\textbf{0}, \textbf{$I_2$})}
    \label{fig:h_error_dw}
\end{figure}

\begin{figure}[!ht]
    \centering
    \begin{subfigure}{0.31\textwidth}
        \includegraphics[width=\textwidth]{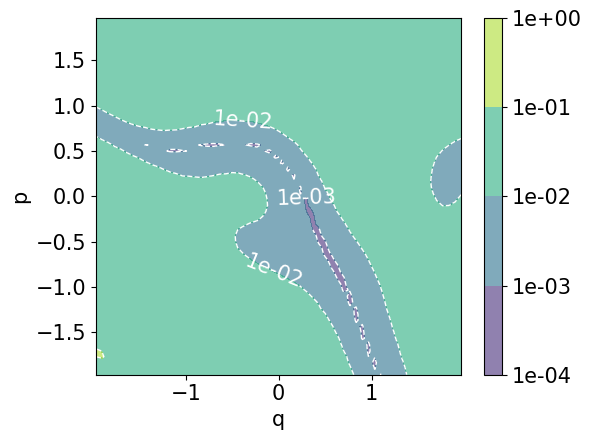}
        \caption{}
        \label{fig:cho_uniform}
    \end{subfigure}
    \hfill
    \begin{subfigure}{0.31\textwidth}
        \includegraphics[width=\textwidth]{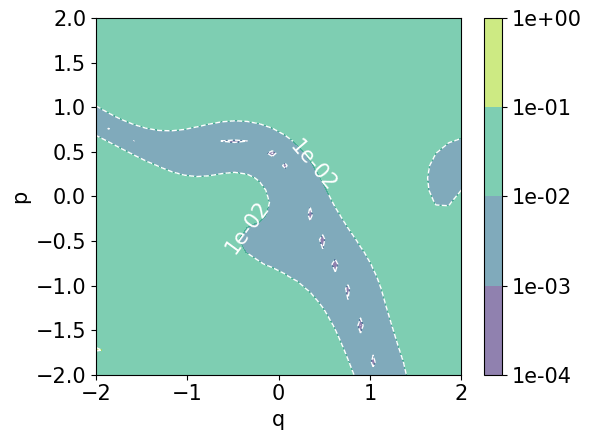}
        \caption{}
        \label{fig:cho_grid}
    \end{subfigure}
    \hfill
    \begin{subfigure}{0.31\textwidth}
        \includegraphics[width=\textwidth]{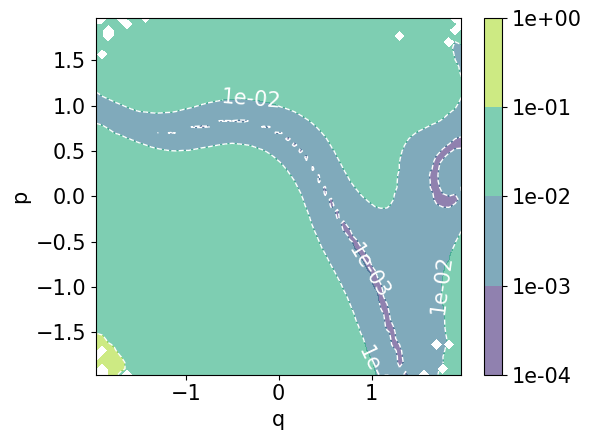}
        \caption{}
        \label{fig:cho_gaussian}
    \end{subfigure}
    \caption{The Hamiltonian prediction error $\|\Hc_{pred} - \Hc_{true}\|_1$ in coupled harmonic oscillator system on test data drawn from 3 different distributions (a)random uniform (b)uniform square grid (c)multivariate gaussian \Nc(\textbf{0}, \textbf{$I_2$})}
    \label{fig:h_error_cho}
\end{figure}

\begin{figure}[!ht]
    \centering
    \begin{subfigure}{0.31\textwidth}
        \includegraphics[width=\textwidth]{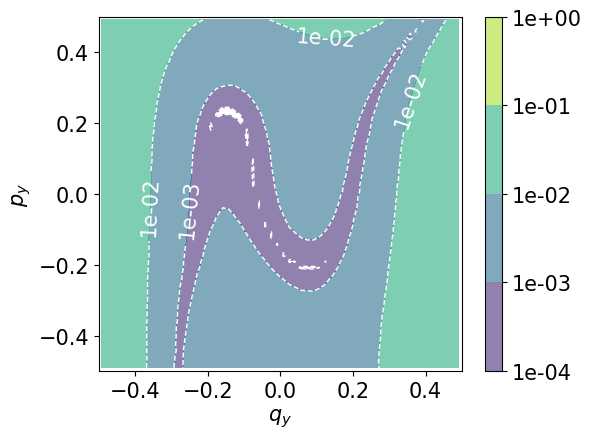}
        \caption{}
        \label{fig:hh_uniform}
    \end{subfigure}
    \hfill
    \begin{subfigure}{0.31\textwidth}
        \includegraphics[width=\textwidth]{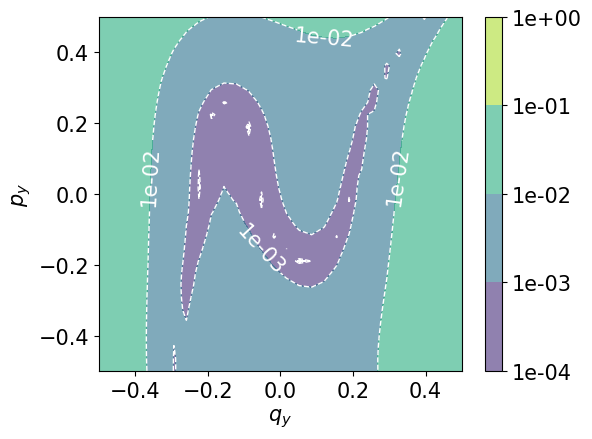}
        \caption{}
        \label{fig:hh_grid}
    \end{subfigure}
    \hfill
    \begin{subfigure}{0.31\textwidth}
        \includegraphics[width=\textwidth]{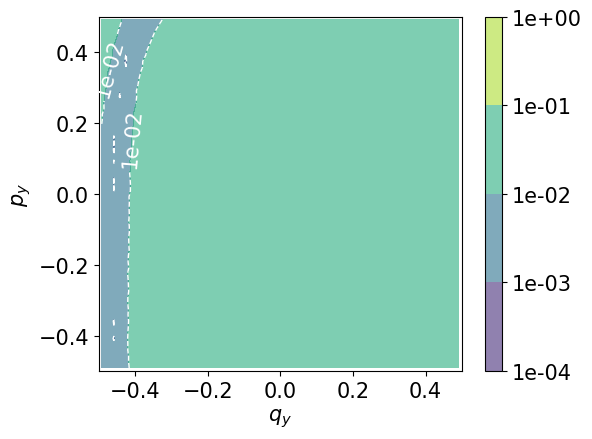}
        \caption{}
        \label{fig:hh_gaussian}
    \end{subfigure}
    \caption{The Hamiltonian prediction error $\|\Hc_{pred} - \Hc_{true}\|_1$ in Henon-Heiles system on test data drawn from 3 different distributions (a)random uniform (b)uniform square grid (c)multivariate gaussian \Nc(\textbf{0}, \textbf{$I_2$})}
    \label{fig:h_error_hh}
\end{figure}

The results for Hamiltonian prediction error $\|\Hc_{pred} -\Hc_{true}\|_1$ over test domain are plotted in figures \ref{fig:h_error_dw}, \ref{fig:h_error_cho} and \ref{fig:h_error_hh} for the three Hamiltonian systems under test respectively. Table~\ref{tab:baseline_comparison} provides a direct comparison of our method against two of the chosen baselines in terms of mean Hamiltonian prediction error over test domain. 


\subsection{Hamiltonian Systems and Results}
\begin{figure}[!ht]
    \centering
    \begin{subfigure}{0.22\textwidth}
        \includegraphics[width=\textwidth]{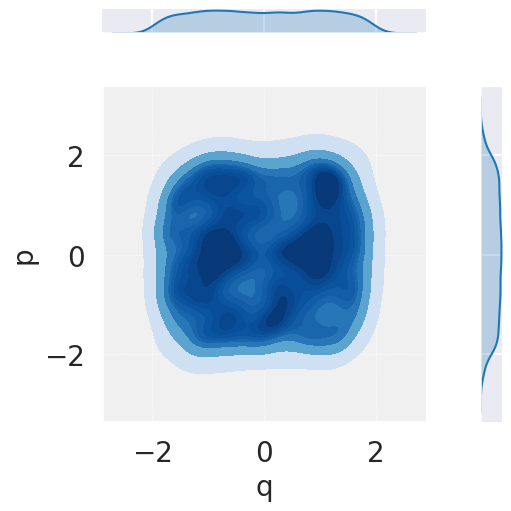}
        \caption{}
        \label{fig:ip_data_dw}
    \end{subfigure}
    \hfill
    \begin{subfigure}{0.22\textwidth}
        \includegraphics[width=\textwidth]{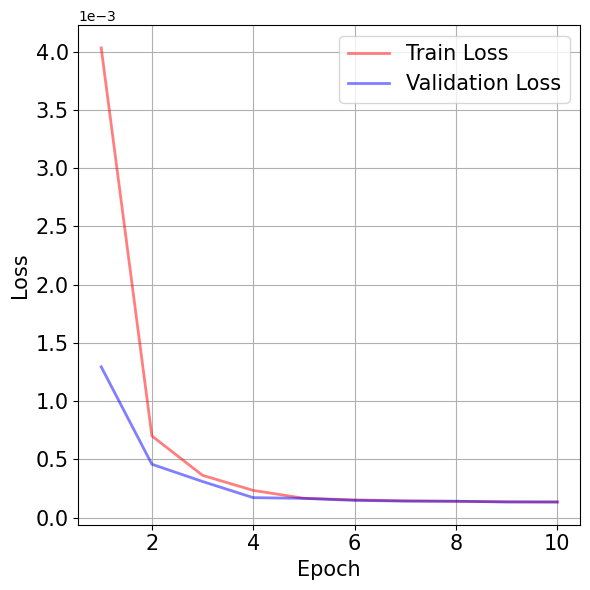}
        \caption{}
        \label{fig:H_error_dw}
    \end{subfigure}\hfill
    \begin{subfigure}{0.22\textwidth}
        \includegraphics[width=\textwidth]{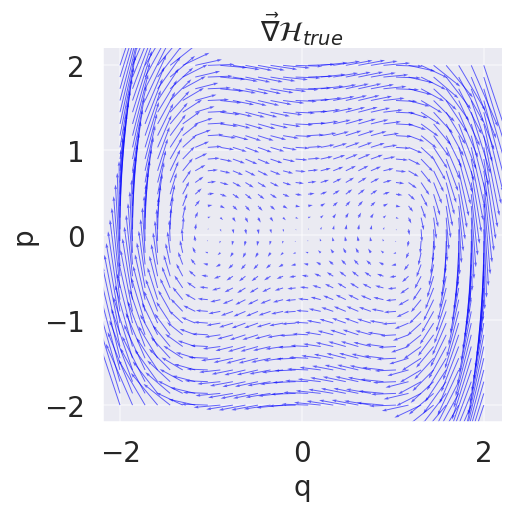}
        \caption{}
        \label{fig:traj_true_dw}
    \end{subfigure}
    \hfill
    \begin{subfigure}{0.22\textwidth}
        \includegraphics[width=\textwidth]{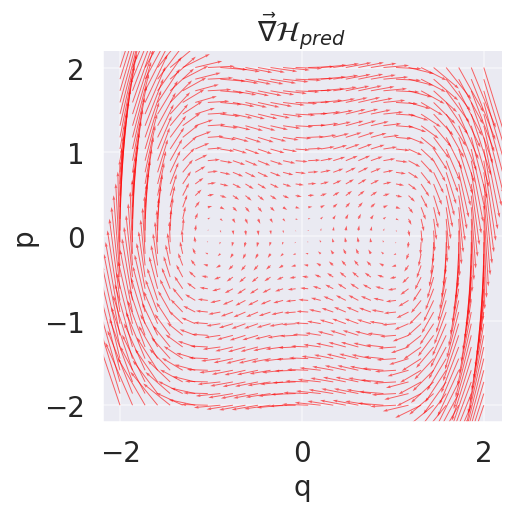}
        \caption{}
        \label{fig:traj_pred_dw}
    \end{subfigure}
    \caption{Representative plots for (a) distribution of training data (b) train and val loss (c) true dynamics (d) predicted dynamics for the double well potential.}
    \label{fig:double_well}
\end{figure}

\textbf{System 1: Double Well potential}\\
Particle in a double-well potential is another commonly studied system in classical and quantum mechanics where where the system has 2 stable fixed points, in our case we consider a symmetrical double potential well with Hamiltonian and governing equations given as:
\begin{equation}\label{eq:state_dw}
\begin{aligned}
\Hc &= \frac{p^2}{2} + \frac{q^4}{4} - \frac{q^2}{2} \\
 \dot{q}&= p,\quad
 \dot{p}= q - q^3.
 \end{aligned}
\end{equation}
 Plots in figure \ref{fig:double_well} show the input data distribution, train and validation loss, and the predicted dynamics for double well system.

\begin{figure}[!ht]
    \centering
    \begin{subfigure}{0.22\textwidth}
        \includegraphics[width=\textwidth]{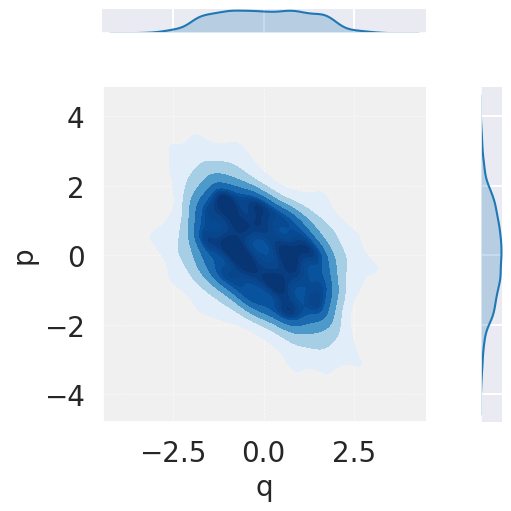}
        \caption{}
        \label{fig:ip_data_co}
    \end{subfigure}
    \hfill
    \begin{subfigure}{0.22\textwidth}
        \includegraphics[width=\textwidth]{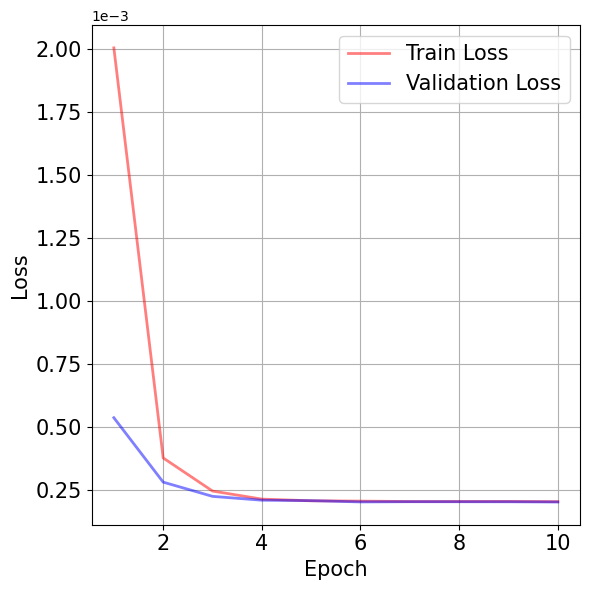}
        \caption{}
        \label{fig:H_error_co}
    \end{subfigure}\hfill
    \begin{subfigure}{0.22\textwidth}
        \includegraphics[width=\textwidth]{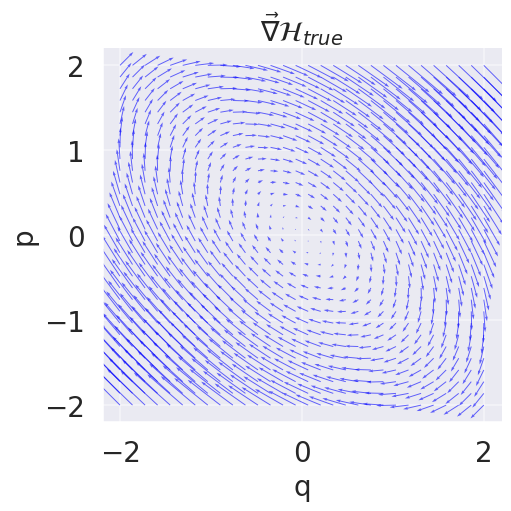}
        \caption{}
        \label{fig:true_traj_co}
    \end{subfigure}
    \hfill
    \begin{subfigure}{0.22\textwidth}
        \includegraphics[width=\textwidth]{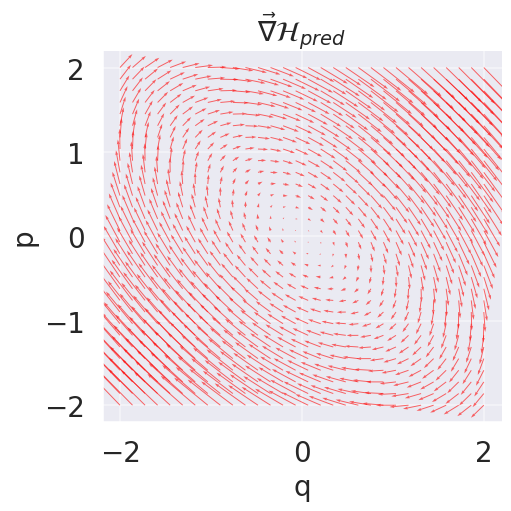}
        \caption{}
        \label{fig:pred_traj_co}
    \end{subfigure}
    \caption{Representative plots for (a) distribution of training data (b) train and val loss (c) true dynamics (d) predicted dynamics for the coupled harmonic oscillator.}
    \label{fig:coupled_ho}
\end{figure}

\textbf{System 2: Coupled Harmonic Oscillator}\\
A coupled Harmonic Oscillator is a simple 1-D non-separable Hamiltonian system with Hamiltonian and governing dynamics given by
\begin{equation}\label{eq:state_coupled_ho}
\begin{aligned}
    \Hc =& \frac{p^2}{2} + \frac{q^2}{2} + \alpha pq\\
    \quad  \dot{q}=& p + \alpha q, \quad  \dot{p}= -(q + \alpha p).
\end{aligned}
\end{equation}
Plots in figure \ref{fig:coupled_ho} show the input data distribution, train and validation loss, and the predicted dynamics for the coupled oscillator system.

\textbf{System 3: Henon-Hieles Potential}\\
We now explore higher-dimensional systems where chaos can emerge. Chaotic systems are deterministic yet unpredictable over long timescales due to exponential error growth, governed by the Lyapunov exponent. However, since these systems follow well-defined Hamiltonians, their dynamics can still be learned from limited observations.
A key example is the Hénon–Heiles (HH) system, a non-separable Hamiltonian model describing a star’s planar motion around a galactic center. While the system exhibits chaotic behavior, stable regions exist, aiding in learning its governing dynamics \cite{barrio2020distribution}. The hamiltonian $\mathcal{H}$ and the corresponding equations of motion are given by:  
\begin{equation}\label{eq:state_hh}
\begin{array}{ll}
\mathcal{H} = \frac{p_x^2 + p_y^2}{2} + \frac{q_x^2 + q_y^2}{2} + q_x^2q_y -\frac{q_y^3}{3}\\
 \dot{q_x} = p_x,\quad
 \dot{q_y} = p_y, \\
 \dot{p_x} = -q_x - 2q_xq_y,\quad
 \dot{p_y} = -q_y - q_x^2 + q_y^2.\\

 \end{array}
\end{equation}

\begin{figure}[!ht]
    \centering
    \begin{subfigure}{0.22\textwidth}
        \includegraphics[width=\textwidth]{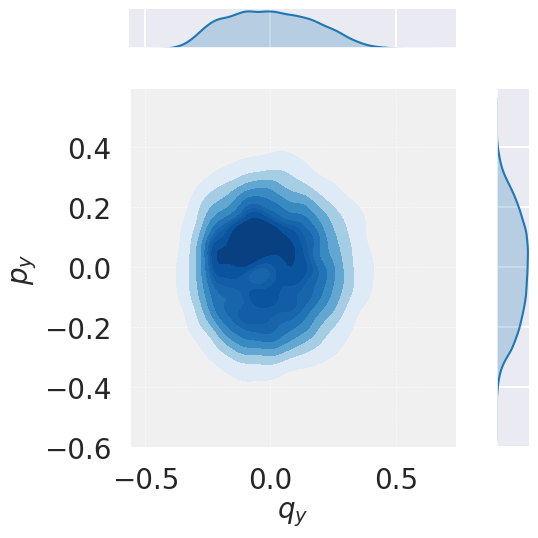}
        \caption{}
        \label{fig:ip_data_hh}
    \end{subfigure}
    \hfill
    \begin{subfigure}{0.22\textwidth}
        \includegraphics[width=\textwidth]{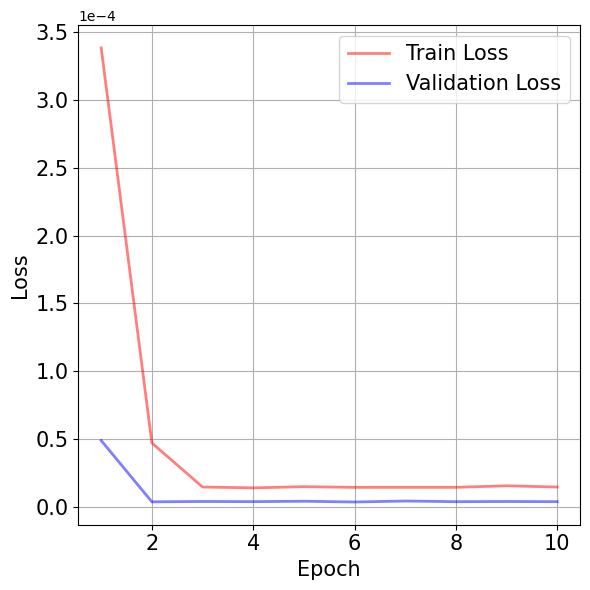}
        \caption{}
        \label{fig:H_error_hh}
    \end{subfigure}\hfill
    \begin{subfigure}{0.22\textwidth}
        \includegraphics[width=\textwidth]{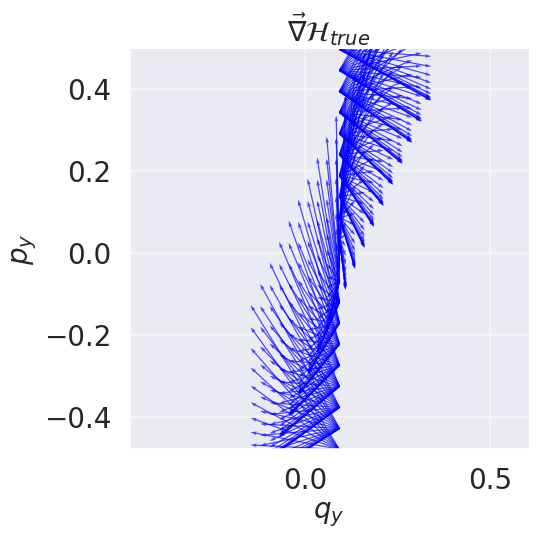}
        \caption{}
        \label{fig:true_traj_hh}
    \end{subfigure}
    \hfill
    \begin{subfigure}{0.22\textwidth}
        \includegraphics[width=\textwidth]{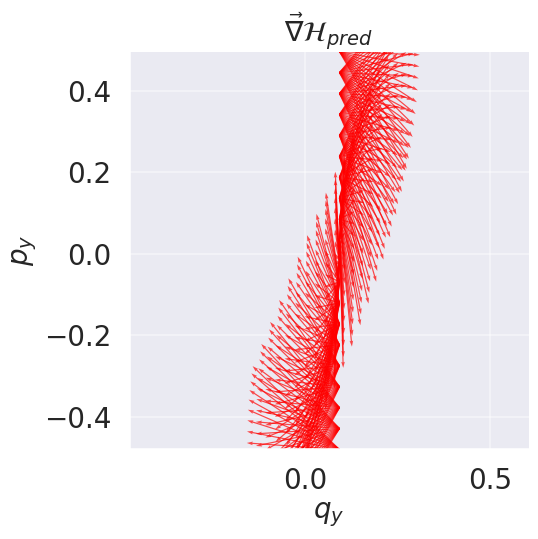}
        \caption{}
        \label{fig:pred_traj_hh}
    \end{subfigure}
    \caption{Representative plots for (a) distribution of training data (b) train and val loss (c) true dynamics (d) predicted dynamics for the Henon-Heiles system. Note that x and y axes here represent projections of y-coordinate of position and momentum for fixed $(p_x, q_x)$}
    \label{fig:henon_heiles}
\end{figure}

The figure \ref{fig:henon_heiles} show the input data distribution, train and validation loss, and the predicted dynamics for Henon-Heiles system.

\subsection{Runtime and Memory performance}\label{ss:mem_perform}
\begin{figure}[!ht]
    \centering
    \begin{subfigure}{0.45\textwidth}
        \includegraphics[width=\textwidth]{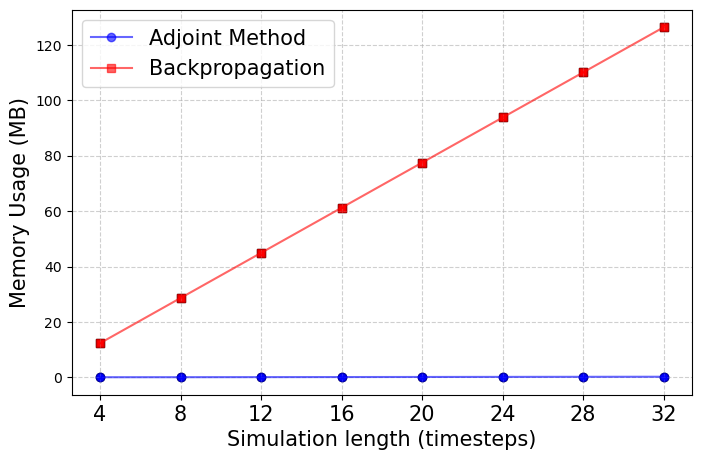}
        \caption{}
        \label{fig:subfig1_memory}
    \end{subfigure}
    \hfill
    \begin{subfigure}{0.45\textwidth}
        \includegraphics[width=\textwidth]{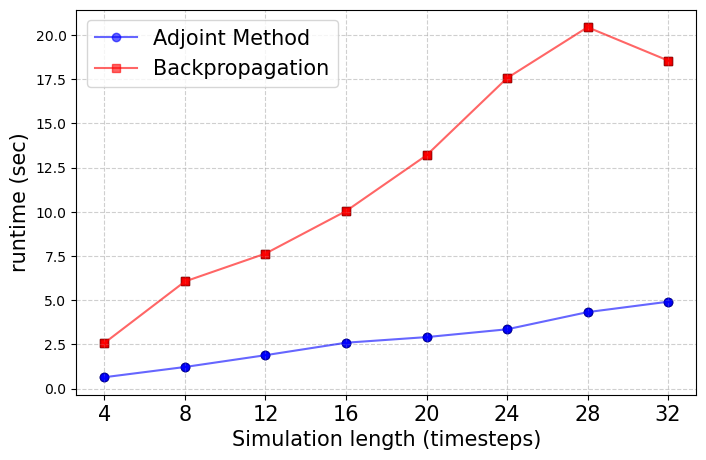}
        \caption{}
        \label{fig:subfig2_runtime}
    \end{subfigure}
    
    \caption{Comparison of (a) memory and (b) runtime profiles for adjoint and backdrop-based gradient evaluation. Each data point corresponds to the metrics evaluated for a single training iteration for a single batch of size 512 for the coupled harmonic oscillator system where the $x$-axis represents number of simulation timesteps of dt=0.01 (Note that as the problem becomes larger the runtime for adjoint state surpasses the backpropagation as it involves solving terminal value problem in the backward pass.)}
    \label{fig:adj_vs_bprop}
\end{figure}

\vspace{0.5em}
In our study, we systematically analyzed the runtime and memory consumption of the adjoint method and backpropagation across increasing simulation lengths, ranging from 4 to 32 time steps. The evaluation was conducted on a benchmark problem involving the training of a coupled harmonic oscillator system with a single batch of size 512. The results as shown in figure \ref{fig:adj_vs_bprop} show a stark contrast in memory scalability between the two approaches: while the adjoint method maintains a constant memory footprint irrespective of the simulation length, the memory usage in backpropagation exhibits a linear growth pattern. This discrepancy arises due to the fundamental difference in how gradients are computed. Backpropagation explicitly stores intermediate states for every time step, whereas the adjoint method reconstructs gradients via a reverse-time integration of the system dynamics, circumventing the need for extensive memory allocation. Interestingly, our runtime analysis also favors the adjoint method for smaller-scale problems, where it demonstrates superior computational efficiency, surpassing backpropagation in execution speed.

\begin{table}[ht]
  \centering
  \scriptsize
  \setlength{\tabcolsep}{4pt}
  \caption{Comparison of the mean absolute error in Hamiltonian prediction 
$\mathbb{E}_{\Omega}[\|\Hc_{\mathrm{pred}} - \Hc_{\mathrm{true}}\|_1]$, 
where $\Omega$ denotes the phase-space test domain. 
Test data are sampled from three different distributions within the domain: 
\textbf{RU}: $(\tilde{\bq},\tilde{\bp})\sim\mathcal{U}([-L,L]^2)$ (uniform over a square domain), 
\textbf{SG}: uniform Cartesian grid on $[-L,L]^2$, and 
\(\boldsymbol{\mathcal{N}}\): $(\tilde{\bq},\tilde{\bp})\sim\mathcal{N}(0,\sigma^2 I_2)$ 
(isotropic Gaussian with $\sigma=L$). 
Here, \emph{Init.} indicates the distribution used to generate the test set.}
  \label{tab:baseline_comparison}
  \begin{tabular}{@{}llccc@{}}
    \toprule
    System & Init. & NSSNN \cite{xiong2020nonseparable} & SHNN \cite{david2023symplectic} & This Work \\
    \midrule
    \multirow{3}{*}{DW} 
      & RU & $7.80 \pm 0.02$ & $0.40 \pm 0.001$ & $1.00\times10^{-2} \pm 1.00\times10^{-5}$ \\
      & SG & $8.00 \pm 1.23$ & $0.41 \pm 0.55$  & $1.2\times10^{-2} \pm 2.6\times10^{-3}$ \\
      & $\mathcal{N}$ & $4.81 \pm 0.02$ & $0.62 \pm 0.005$ & $1.57\times10^{-1} \pm 3.3\times10^{-3}$ \\
    \midrule 
    \multirow{3}{*}{CHO} 
      & RU & $0.80 \pm 0.002$ & $1.3\times10^{-2} \pm 3\times10^{-5}$ & $2.5\times10^{-2} \pm 6\times10^{-5}$ \\
      & SG & $0.86 \pm 0.14$  & $1.3\times10^{-2} \pm 1.68\times10^{-3}$ & $2.6\times10^{-2} \pm 3.3\times10^{-3}$ \\
      & $\mathcal{N}$ & $0.79 \pm 0.003$ & $1.7\times10^{-2} \pm 1.6\times10^{-4}$ & $2.8\times10^{-2} \pm 2.7\times10^{-4}$ \\
    \midrule 
    \multirow{3}{*}{HH} 
      & RU & $5.55\times10^{-1} \pm 2.63\times10^{-2}$ & $1.51\times10^{-2} \pm 4.42\times10^{-4}$ & $1.14\times10^{-2} \pm 4.26\times10^{-4}$ \\
      & SG & $5.03\times10^{-1} \pm 2.8\times10^{-3}$  & $1.38\times10^{-2} \pm 5.38\times10^{-2}$ & $1.02\times10^{-2} \pm 5.12\times10^{-5}$ \\
      & $\mathcal{N}$ & $1.65 \pm 9.61\times10^{-3}$ & $1.03\times10^{-1} \pm 7.79\times10^{-4}$ & $9.11\times10^{-2} \pm 7.69\times10^{-4}$ \\
    \bottomrule 
  \end{tabular}
\end{table}

\section{Conclusion}

The adjoint approach normally results in gradients that differ from backpropagation, unless the adjoint system is computed using the cotangent lift of the numerical integrator used in the forward propagation, in which case the adjoint approach yields gradients that coincide with backpropagation. More generally, this holds when the adjoint system is integrated using a symplectic partitioned integrator, which recovers the numerical integrator for the forward propagation when restricted to the forward flow.

When the forward flow is Hamiltonian, as is the case for Hamiltonian Neural Networks, it is natural to discretize the forward flow using a symplectic integrator. Using the same symplectic integrator on the adjoint variables will lead to a discretization of the adjoint system that yields gradients that also coincide with backpropagation, leading to an efficient method for training Hamiltonian Neural Networks.

In our work, we adopt such an approach, using implicit symplectic partitioned Runge--Kutta methods. Symplectic methods are generally implicit for non-separable Hamiltonians, unless one artificially doubles the number of variables in an augmented formulation. However, contrary to conventional wisdom, implicit SPRK methods can be very efficiently implemented by using an explicit RK method of the same order as a predictor, and using a few fixed-point iterations of the SPRK method as a corrector. Therefore, Hamiltonian Neural Networks with a non-separable Hamiltonian ansatz can be efficiently trained using implicit SPRK discretization by applying the adjoint method combined with the predictor-corrector fixed-point iteration.

\subsection*{Disclaimer}
This paper was prepared for information purposes
and is not a product of HSBC Bank Plc. or its affiliates.
Neither HSBC Bank Plc. nor any of its affiliates make
any explicit or implied representation or warranty and
none of them accept any liability in connection with
this paper, including, but not limited to, the completeness,
accuracy, reliability of information contained herein and
the potential legal, compliance, tax or accounting effects
thereof. Copyright HSBC Group 2024.

\subsection*{Data Availability}
The supplementary material for reproducibility of results can be found \href{https://github.com/choudharyharsh122/HNN/}{here}

\section{Acknowledgements}
\begin{itemize}
\item HC and VK acknowledge support from the Czech National Science Foundation under Project 24-11664S

\item The computational resources used in this work were supported by the project RCI (Reg. No. CZ.02.1.01/0.0/0.0/16\_019/0000765), funded by the European Union. The project was active from March 1, 2018, to October 31, 2022.
\end{itemize}



\bigskip
\bigskip
\bibliography{sn-bibliography}

\end{document}